\pdfoutput=1

\documentclass[11pt]{article}

\usepackage[final]{acl}

\usepackage{times}
\usepackage{latexsym}
\usepackage{amssymb}

\usepackage[T1]{fontenc}

\usepackage[utf8]{inputenc}

\usepackage{microtype}

\usepackage{inconsolata}

\usepackage{graphicx}

\usepackage[shortlabels,inline]{enumitem}
\usepackage{tikz,lipsum}
\usepackage[most]{tcolorbox}
\usepackage{ragged2e}
\usepackage[dvipsnames]{xcolor}
\usepackage{amsmath}
\usepackage{booktabs}
\usepackage{tabularray}
\usepackage{arydshln}
\usepackage{stmaryrd}
\usepackage{marvosym}
\usepackage{colortbl}
\usepackage{multicol}
\usepackage{multirow}
\usepackage{float}

\usepackage{makecell}
\usepackage{cleveref}
\crefname{section}{\S}{\S}
\crefname{table}{Table}{Tables}
\crefname{figure}{Fig.}{Figs.}
\crefname{algorithm}{Alg.}{}
\crefname{ALC@unique}{Line}{Lines}
\crefname{equation}{Eq.}{Eqs.}
\crefname{appendix}{App.}{Apps.}
\crefformat{section}{\S#2#1#3} 
\usepackage{soul}
\definecolor{tablegray}{RGB}{223, 242, 252}

\usepackage{todonotes}

\usepackage[shortlabels,inline]{enumitem}
\usepackage{tikz,lipsum}
\usepackage[most]{tcolorbox}
\usepackage{ragged2e}
\usepackage[dvipsnames]{xcolor}
\usepackage{amsmath}
\usepackage{booktabs}
\usepackage{tabularray}
\usepackage{arydshln}
\usepackage{stmaryrd}
\usepackage{marvosym}
\usepackage{colortbl}
\usepackage{multicol}
\usepackage{multirow}
\usepackage{float}

\usepackage{cleveref}
\crefname{section}{\S}{\S}
\crefname{table}{Table}{Tables}
\crefname{figure}{Fig.}{Figs.}
\crefname{algorithm}{Alg.}{}
\crefname{ALC@unique}{Line}{Lines}
\crefname{equation}{Eq.}{Eqs.}
\crefname{appendix}{App.}{Apps.}
\crefformat{section}{\S#2#1#3} 
\usepackage{soul}
\definecolor{tablegray}{RGB}{223, 242, 252}

\usepackage{todonotes}

\newcommand{\scng}[1]{%
  \ifdim #1 pt > 0 pt%
    \scriptsize{\textcolor{ForestGreen}{(+#1)}}
  \else%
    \scriptsize{\textcolor{red}{(-#1)}}
  \fi%
}
\newcommand{\scngg}[1]{%
  \ifdim #1 pt > 0 pt%
    \scriptsize{\textcolor{BurntOrange}{(+#1)}}
  \else%
    \scriptsize{\textcolor{YellowOrange}{(-#1)}}
  \fi%
}

\usepackage{xspace}
\newcommand{\ours}{\textbf{\texttt{Q\textsubscript{2}E}}\xspace}

\title{\ours: \underline{Q}uery-to-\underline{E}vent Decomposition for Zero-Shot Multilingual Text-to-Video Retrieval}

\newcommand{\DataReleaseURL}[0]{\url{https://dipta007.github.io/Q2E/}}

\author{
\textbf{Shubhashis Roy Dipta, Francis Ferraro} \\
  Department of Computer Science and Electrical Engineering\\
  University of Maryland Baltimore County\\
  Baltimore, MD 21250 USA \\
  \texttt{\{sroydip1,ferraro\}@umbc.edu} \\
}

\begin{document}
\maketitle

\begin{abstract}
Recent approaches have shown impressive proficiency in extracting and leveraging parametric knowledge from Large-Language Models (LLMs) and Vision-Language Models (VLMs). %
In this work, we consider how we can improve the retrieval of videos related to complex real-world events by automatically extracting latent parametric knowledge about those events.
We present \ours: a  \underline{Q}uery-to-\underline{E}vent decomposition method for zero-shot multilingual text-to-video retrieval, adaptable across datasets, domains, LLMs, or VLMs. Our approach demonstrates that we can enhance the understanding of otherwise overly simplified human queries by decomposing the query using the knowledge embedded in LLMs and VLMs. We additionally show how to apply our approach to both visual and speech-based inputs. To combine this varied multimodal knowledge, we adopt entropy-based fusion scoring for zero-shot fusion. 
\ours outperforms the previous SOTA on the MultiVENT dataset by \textbf{8 NDCG points}, while improving on MSR-VTT and MSVD by \textbf{4 and 3 points}, respectively, outperforming several existing retrieval methods, including many fine-tuned and SOTA zero-shot approaches.
We have released both code and data.\footnote{\DataReleaseURL}

\end{abstract}

\section{Introduction}

\begin{figure}[!t]
    \centering
    \includegraphics[width=\columnwidth]{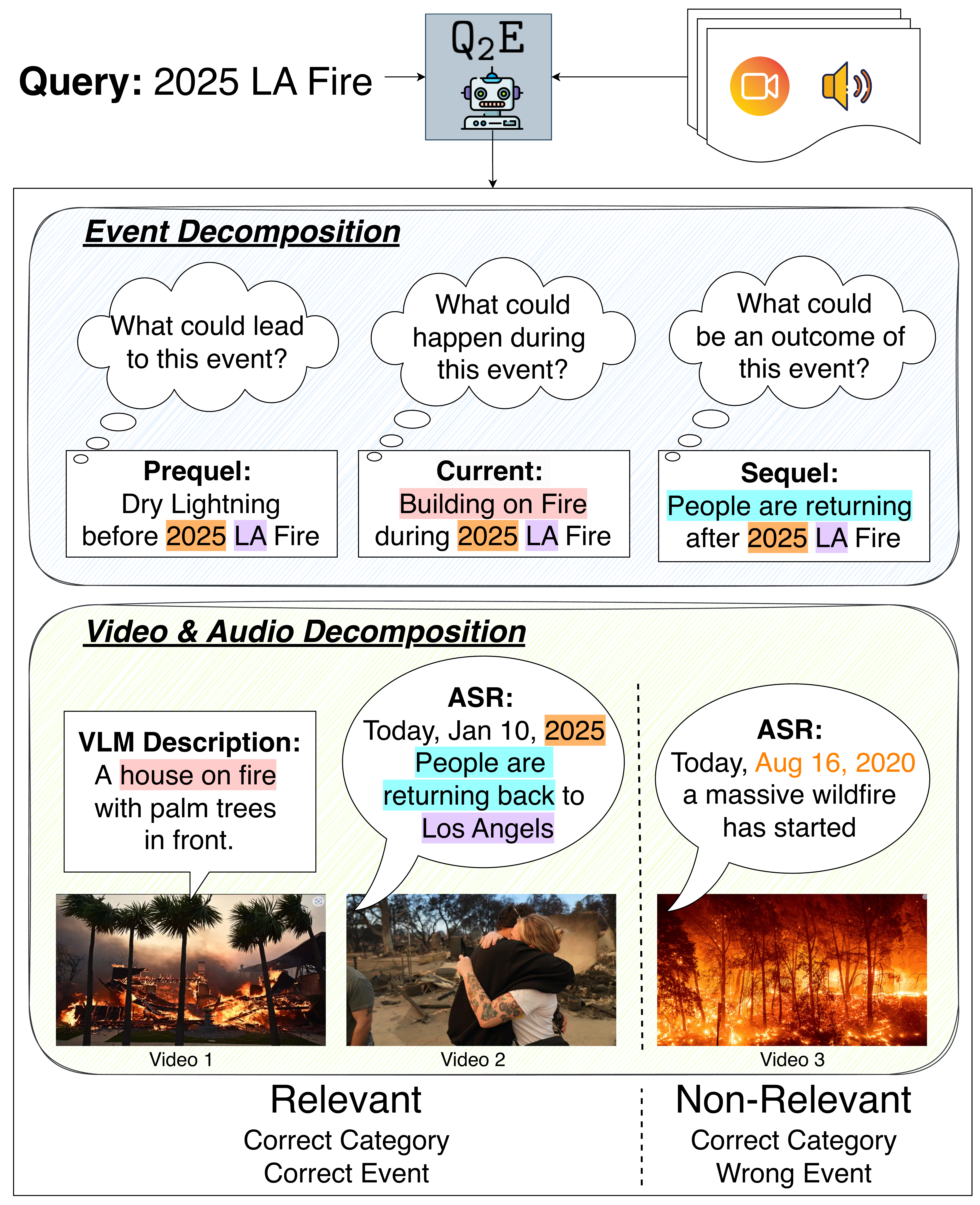}
\caption{
In \ours, we extract prequel, current, and sequel events, along with an audio transcript and video description to enrich the query and video context, respectively. %
This decomposed queries are matched across visual, textual, and speech-based descriptions (matching phrases are highlighted in the same color) %
enabling the retrieval of the correct videos while effectively filtering out a visually similar but non-relevant video.
\label{fig:intro}
}
\vspace{-2mm}
\end{figure}

Making sense of complex real-world events, such as a natural disaster or fire, requires understanding more than just one individual event or action. In addition to the multiple events that occur during that broader event (such as structures burning or evacuation), one might need to understand the events that lead up to it (such as dry conditions), or that arise from it (such as recovery and rebuilding). %
Videos present a rich medium for conveying this multi-faceted information, as they contain much more than what is visible at any given point in time: they can show changes and dynamic actions, and through an audio track, allow for natural language descriptions, either in-the-moment or after-the-fact, about that event. %
Being able to automatically retrieve videos about a queried event could provide the information necessary for users to better make sense of events they need to know about. %
However, there are multiple challenges to this: first, while a query might be a concise reference to a particular event, such as ``2025 LA fire,'' a user would still want all this information. Second, this information might be scattered across multiple videos, with any given video only showing a portion of the information needed. Third, these videos might not be in the language that the user knows. %

We present a novel zero-shot method, \ours, that addresses these problems. It consists of a proposed event decomposition method that \textbf{leverages the embedded knowledge that LLMs have about those {\it \bfseries types} of complex events} and uses this knowledge to \textbf{automatically decompose and elaborate (or expand) a query}. We  show how to use this knowledge to \textbf{identify and score salient visual and multi-lingual audio-based features from the video}, and combine this information to return the sought after videos. %

To achieve this, \ours integrates the pre-existing knowledge of LLMs and VLMs to enhance text-to-video retrieval. %
Underlying our approach are three key insights: %
\begin{enumerate*}[label=\textbf{(\arabic*})]
    \item Transferring knowledge from LLMs through query decomposition enhances the understanding of coarse-grained human queries. This provides richer context and enables the retrieval of videos showing what can lead to or result from the target event that would otherwise be overlooked.
    \item While VLM-based captioning and large ASR models can generally describe an image/video frame and transcribe a portion of speech, the outputs can be repetitive, noisy, missing broader themes that occur across frames. However, a LLM based refiner can help to remove those noises and redundancies for empirical effectiveness.
    \item In our approach, determining whether an individual video is relevant requires aggregating across multiple forms of similarity and relevancy judgments, each of varying strength. Entropy-based rank fusion provides an intuitive way of performing this aggregation and ranking, while surpassing both simple fusion methods (mean, max) and more complex approaches like Reciprocal Rank Fusion.
\end{enumerate*}
Using these insights,  \ours achieves strong performance gains on three varied and challenging datasets~\citep{chen-dolan-2011-collecting,xu2016msr,sanders2023multivent}. %

\textbf{But isn't video retrieval solved?} Widely used video retrieval websites can give the impression that video retrieval is a solved problem: a user types in a query, and gets back a nicely curated list of relevant videos. However, such approaches can leverage rich platform-specific metadata or human-annotated data (like titles or search-optimized descriptions) that go beyond the information conveyed \textit{directly} in that video. %
For example, as shown in \cref{fig:intro}, there are many different videos of \textit{fires}, but if a user provides a basic query such as ``\textit{2025 LA fire}'' in order to find videos providing information about \textit{that} specific fire, how can a system satisfy this request? %
Most current text-to-video retrieval systems are trained and evaluated on classic datasets, such as MSR-VTT \citep{xu2016msr} and MSVD \citep{chen-dolan-2011-collecting}, which feature generic, high-level queries like ``\textit{a person is explaining something}'' (a random query from MSR-VTT), rather than referencing a complex real-world event. %
Recent work~\citep{liu2024unifying} has also found that users typically input basic queries, expecting the system to capture ``fine-grained semantic concepts'' and understand complex relationships between videos and text. %
Finally, \citet{sanders2023multivent} demonstrate the disparity that exists when retrieving non-English videos. %
Our work addresses these challenges by showing how to automatically expand the query with information needed to understand the event, and how to get video+audio features reflective of this information to retrieve highly specific videos across different languages. %
In doing so, we show how the extensive knowledge embedded in Large-Language Models (LLMs) and Vision-Language Models (VLMs) can be effectively transferred to text-to-video retrieval.

In summary, our contributions are:
\begin{itemize}[leftmargin=*]
\setlength\itemsep{0pt}
\item We propose a novel method that leverages the embedded world knowledge of LLMs to enrich human queries with prequel, current, and sequel-based events.
\item We use large video, speech, and language models to capture critical event- and query-relevant information communicated across those multiple modalities, how to assess that information for query relevance, and how to combine those assessments via entropy-based rank fusion.

\item Through extensive ablation studies, we have shown the importance of each component of our method in the final performance.

\item Our zero-shot method improved performance on the widely used text-to-video retrieval dataset, MSR-VTT (\textbf{+4 NDCG}) \& MSVD (\textbf{+3 NDCG}) as well as on the multi-lingual event-focused retrieval dataset, MultiVENT (\textbf{+8 NDCG}).
\end{itemize}

\section{\ours: Query-to-Event Decomposition}

\begin{figure*}[t]
    \centering
    \includegraphics[width=\textwidth]{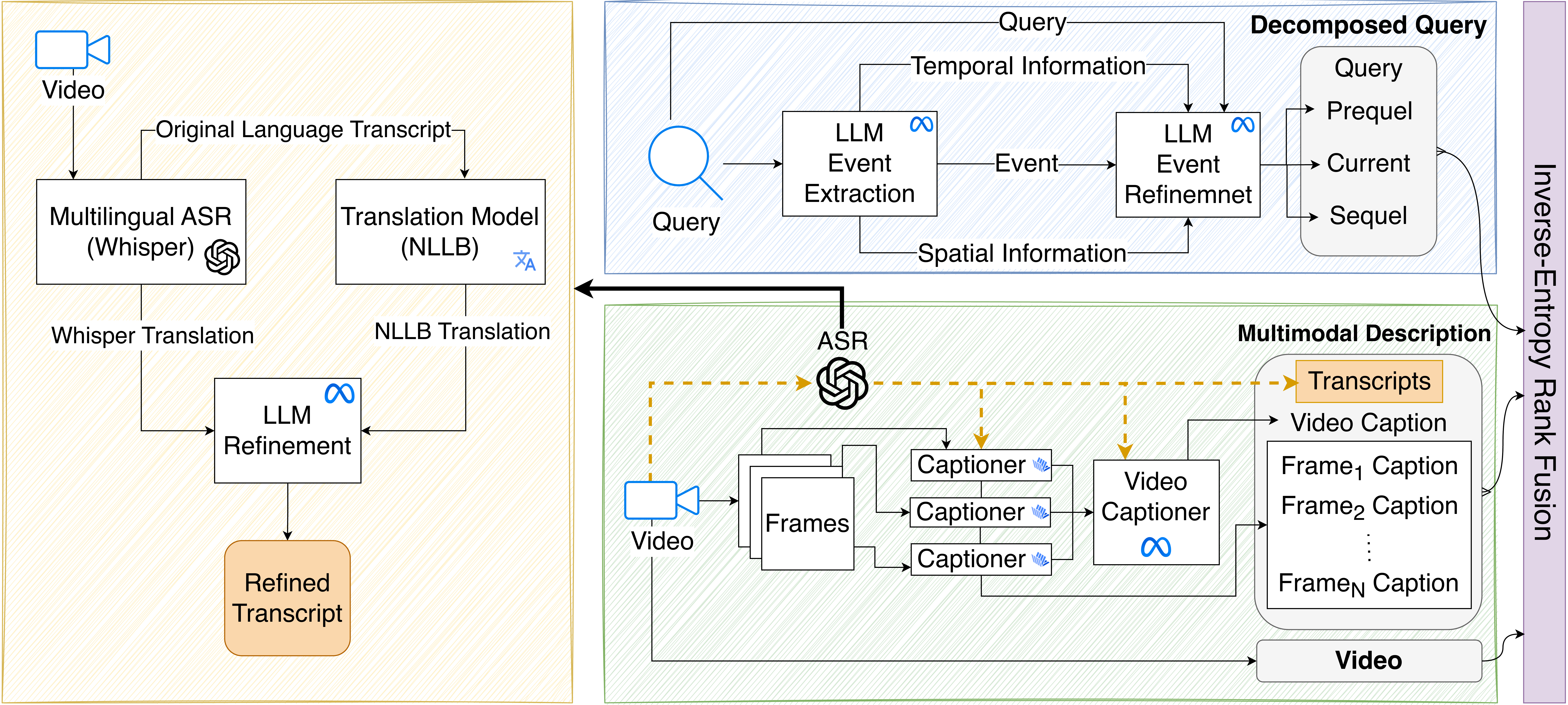}
\caption{ %
\ours, the complete framework of our text-to-video retrieval system. The blue box represents the event decomposition module, while the green box illustrates the video decomposition module. The orange box represents our multi-layer Audio Decomposition module. The purple box fuses the ranks calculated from query-video, query-descriptions, and event-descriptions (event = prequel + current + sequel, description = multimodal description). 
In the non-ASR variant, the components within the orange box and orange dotted lines are excluded.
\label{fig:main_with_asr}
}
\end{figure*}

Modern dense retrieval approaches commonly compute an embedding-based similarity (e.g., cosine) between a query and each candidate; those scores rank candidates, and the top-scoring candidates are returned. %
For example, in video retrieval, one could embed the user's query using an appropriate LLM, embed the video using an appropriate video embedder, and then compute the cosine similarity between those embeddings~\cite{sanders2023multivent}. This ``query vs. video'' similarity score is used to determine how relevant to video is to the query.

\ours (shown in \cref{fig:main_with_asr}) builds upon this general embedding similarity computation. To better capture the diverse and nuanced multimodal information conveyed in videos and human queries---including the event decompositions---we must extend those similarity computations. %
In order to capture the rich event information we get from our decomposition (\cref{sec:query_decom}), a single embedding of the video is not sufficient. We need to additionally get \textit{multimodal descriptions} of the video (\cref{sec:video_decom,sec:asr}).
Using these components, \ours computes five types of scores $S_{i,v}$ for each video $v$:
(a) a query vs. video score, as described above; (b) prequel, (c) current (d) sequel vs. multimodal-description score and (e) a query vs. multimodal-description score.
While (a) helps to identify coarse-grained features such as ``fire'', (b-e) provide fine-grained contextual details. These scores complement one another, which we merge through entropy-based fusion and ranking (\cref{sec:score_fusion}).

\subsection{Event Decomposition} \label{sec:query_decom}
Like any traditional text-video retrieval system, a user provides a natural language query, which can be a couple of words to 10+ words. In \ours, we have used an LLM to decompose it into three sub-events: (1) Prequel, (2) Current, and (3) Sequel. Based on initial experiments, and to manage computation, we only considered five of prequel, current, and sequel events. We provide our template prompts for each in \cref{sec:app_prompts}.
\begin{description}[leftmargin=5mm,itemsep=0pt]
\item[Prequel] \label{sec:prequel}
``Prequel'' refers to events that can lead to the current event described in the query. An event can have multiple prequels; for example, a wildfire can have strong wind, high-temperature alert, and dry lightning as prequels. Prequel events may \textit{enable} the event in the query, but do not necessarily \textit{cause} it. 

\item[Current] \label{sec:current}
``Current'' mainly refers to the simple decomposed events that occur during the event described in the query. This method decomposes the queries into sub-events likely to be observed. E.g., ``\textit{2025 LA Fire}'' can be decomposed into ``Building on Fire'', or ``Smoke everywhere.''

\item[Sequel] \label{sec:sequel}
``Sequel'' events can be the results from the event described in the query. Like the prequel, one event can lead to multiple sequel events.
\end{description}

\paragraph{Decomposition Refinement} \label{sec:decom_refinement}
We refine these event decompositions with temporal, spatial, and primary event data. 
For this, we have used the same LLM to extract the temporal (2025), spatial (LA), and primary event (Fire) information. Although this rich information can be combined templatically, e.g., ``P/S/C (2025, LA, Fire),'' during initial development experiments, we found that refining it to read as a more natural, human-provided query markedly improved performance. We produced this refined query automatically using an LLM; we show this in \cref{app:refine_query}.

\subsection{Video Decomposition} \label{sec:video_decom}
In practice, we found that embeddings of videos / video frames were not sufficient to capture the varied actions described by our event decomposition through basic similarity computations. While alternative similarity approaches, such as cross-encoders, may be effective, they present computational challenges, requiring running the cross-encoder on every video/query pair. To address this, we instead extract captions from the videos themselves, with the aim that these descriptions would be sufficiently similar to our query event decomposition. We found two approaches to be effective for capturing complementary local and global elements across the video: (1) generating contextualized descriptions per selected frame, and (2) generating holistic video descriptions.

\paragraph{Contextualized Frame Descriptions} \label{sec:contextualized_frame_caption}
We uniformly sampled 16 frames from the videos (determined by trading off performance vs. computation--- ablation in \cref{sec:app_num_of_frames}). We have tried uniform sampling and a scene-change-based sampling method \footnote{\url{https://www.scenedetect.com/}} for frame sampling, and found that uniform sampling, despite its simplicity, works better than scene-change-based sampling (ablation in \cref{sec:app_frame_selection_method}). As the video contains temporal information and by design, video frames should be connected semantically, so rather than captioning each frame in isolation, we used a sliding window (window size=2) approach and provided the VLM model with the previous frame’s caption (when available). %
Conditioning on previous frame's captions helped contextualize subsequent captions. Per video, we have K contextualized frame captions. %

\paragraph{Video Descriptions} \label{sec:video_caption}
We found that current VLM models can perform dense captioning but miss the ``bigger picture.'' For instance, a VLM might describe a video of a wildfire with \textit{people running} and \textit{fire everywhere} but fail to recognize the higher-level conclusion of ``wildfire.'' %
To address this limitation, we provided an LLM model with all 16-frame descriptions from above and asked it to summarize them into a single video-dense caption, preserving the order of the frame captions.
We carefully designed the prompt (\cref{sec:video_captioner}) such that it preserves temporal information and focuses on broader picture rather than color, shape or objects.

\subsection{Audio Decomposition}
\label{sec:asr}
Speech is a critical component of many videos, especially when visual elements alone may not be sufficient enough to convey the full context. %
Despite using a multilingual ASR model~\citep{radford2023robust}, the multilingual videos posed a unique challenge to get consistent and accurate ASR transcriptions due to the background noise and different dialects of the real life videos in MultiVENT. %
To address this, we propose a multi-layer translation pipeline (\cref{fig:main_with_asr}, left orange box). This pipeline leverages three models: 
(1) a multilingual ASR agent, %
\texttt{whisper-v3}~\citep{radford2023robust}, which transcribes audio into both the original language and its English translation; 
2) a translator agent, %
\texttt{NLLB}~\citep{costa2022no} to generate English transcriptions by translating the original transcript; and 
(3) a refiner agent, %
\texttt{Llama-70B}~\citep{dubey2024llama}, which refines both the English translations. Results showed all three are critical for sufficient transcription and translation quality (\cref{sec:app_diff_layer_translation}).

\subsection{Scoring and Score Fusion} \label{sec:score_fusion}

\paragraph{Query vs. Video}
We utilized the original query and video to compute a similarity score. To do this, we follow \citet{sanders2023multivent} in adapting an image encoder: we uniformly sample and embed 16 frames, averaging their embeddings to generate the final video embedding. Cosine similarity was applied to compute the similarity score, which was then scaled to the 0–100 range to ensure comparability with other scores:
$ %
    S_{i,v} | Q = 100.0 \times \operatorname{Sim_v}(Q, V), %
$ %
where $S_{i,v} | Q$ denotes the score vector for a given query $Q$, and $\operatorname{Sim_v}(x, y)$ returns the similarity score between query $x$ and video $y$ using an embedding model. %
Primarily we have used MultiCLIP \citep{delitzas2023multi} due to its superior performance with multilingual videos compared to the CLIP \citep{radford2021learning} model, although we show in our experiments that our approach successfully uses other image encoders too.

\begin{table*}[!ht]
\resizebox{\textwidth}{!}{
    \begin{tabular}{llccccccccccr}
    \toprule
    Model & \thead{Knowledge\\Augmentation} & R@1$\uparrow$ & R@5$\uparrow$ & R@10$\uparrow$ & P@1$\uparrow$ & P@5$\uparrow$ & P@10$\uparrow$ & MRR$\uparrow$ & NDCG$\uparrow$ & MAP$\uparrow$ & MnR$\downarrow$ & MdR$\downarrow$ \\
    \midrule
    
    & & \multicolumn{11}{c}{\textbf{MultiVENT}} \\
    \cmidrule(ll){3-13}
    MultiCLIP & - & 9.83 & 44.32 & 70.82 & 88.80 & 80.77 & 65.25 & 0.92 & 75.34 & 86.33 & 22.12 & \textbf{6} \\
    MultiCLIP & Event & 10.24 & 46.71 & 75.76 & 92.28 & 85.25 & 69.73 & \textbf{0.95} & 80.04 & 89.42 & 21.13 & \textbf{6} \\    

    \rowcolor{tablegray}
    MultiCLIP & ASR + Event & \textbf{10.32} & \textbf{49.00} & \textbf{79.60} & \textbf{93.05} & \textbf{88.73} & \textbf{73.09} & \textbf{0.95} & \textbf{83.24} & \textbf{91.20} & \textbf{16.44} & \textbf{6} \\

    \addlinespace[1mm]
    \addlinespace[1mm]
    
    InternVideo2-1B & - & 5.60 & 28.92 & 49.12 & 51.35 & 53.28 & 45.44 & 0.68 & 50.43 & 63.77 & 235.49 & 11 \\
    InternVideo2-1B & Event & 9.54 & 40.88 & 63.40 & 87.26 & 75.21 & 58.73 & 0.92 & 69.15 & 83.96 & 53.84 & 7 \\    
    
    \rowcolor{tablegray}
    InternVideo2-1B & ASR + Event & \textbf{10.24} & \textbf{44.94} & \textbf{70.79} & \textbf{92.28} & \textbf{81.85} & \textbf{65.14} & \textbf{0.95} & \textbf{76.10} & \textbf{88.09} & \textbf{42.81} & \textbf{6} \\
    
    \addlinespace[1mm]

    & & \multicolumn{11}{c}{\textbf{MSR-VTT-1kA}} \\
    \cmidrule(ll){3-13}
    MultiCLIP & - & 43.52 & 69.05 & 76.88 & 43.62 & 13.85 & 7.71 & 0.54 & 59.72 & 54.27 & 20.29 & 2 \\
    MultiCLIP & Event & 44.52 & 71.26 & 79.40 & 44.62 & 14.29 & 7.96 & 0.56 & 61.51 & 55.84 & \textbf{17.91} & 2 \\
    
    \rowcolor{tablegray}
    MultiCLIP & ASR + Event & \textbf{46.23} & \textbf{73.37} & \textbf{81.71} & \textbf{46.33} & \textbf{14.71} & \textbf{8.19} & \textbf{0.58} & \textbf{63.59} & \textbf{57.83} & 18.73 & 2 \\

    \addlinespace[1mm]
    \addlinespace[1mm]

    InternVideo2-1B & - & 52.56 & 73.07 & 80.10 & 52.66 & 14.65 & 8.03 & 0.62 & 66.07 & 61.62 & 25.12 & \textbf{1} \\
    InternVideo2-1B & Event & 53.47 & 73.57 & 82.11 & 53.57 & 14.75 & 8.23 & 0.62 & 67.16 & 62.47 & 16.73 & \textbf{1} \\
    
    \rowcolor{tablegray}
    InternVideo2-1B & ASR + Event & \textbf{56.28} & \textbf{76.58} & \textbf{83.72} & \textbf{56.38} & \textbf{15.36} & \textbf{8.39} & \textbf{0.65} & \textbf{69.53} & \textbf{65.06} & \textbf{15.85} & \textbf{1} \\

    & & \multicolumn{11}{c}{\textbf{MSVD}} \\
    \cmidrule(ll){3-13}
    MultiCLIP & - & 54.91 & 80.82 & 87.18 & 56.24 & 16.96 & 9.23 & 0.67 & 71.69 & 66.79 & 11.46 & 1 \\
    
    \rowcolor{tablegray}
    MultiCLIP & Event & \textbf{57.29} & \textbf{83.26} & \textbf{89.18} & \textbf{58.68} & \textbf{17.48} & \textbf{9.45} & \textbf{0.70} & \textbf{74.10} & \textbf{69.29} & \textbf{9.56} & \textbf{1} \\
    
    \addlinespace[1mm]
    \addlinespace[1mm]
    
    InternVideo2-1B & - & 63.59 & 84.85 & 89.46 & 65.03 & 17.85 & 9.49 & \textbf{0.74} & 77.51 & 73.68 & 12.32 & \textbf{1} \\
    
    \rowcolor{tablegray}
    InternVideo2-1B & Event & \textbf{63.77} & \textbf{85.30} & \textbf{89.99} & \textbf{65.24} & \textbf{17.94} & \textbf{9.54} & \textbf{0.74} & \textbf{77.84} & \textbf{73.95} & \textbf{9.40} & \textbf{1} \\
    
    \bottomrule
    \end{tabular}
}

\caption{Performance comparison on  MultiVENT, MSRVTT ``1k-A'' and MSVD dataset. Our method with \ours, even without ASR, improves NDCG by 5–19 points on the MultiVENT dataset, 1-2 points on the MSRVTT dataset and 1-3 points on the MSVD dataset. Adding ASR further enhances performance, with an additional 3–7 points on MultiVENT and 2 points on MSRVTT, resulting in a total improvement of \textbf{8–26 points} on MultiVENT and \textbf{3–4 points} on MSRVTT, across two different encoders. Since the original MSVD videos are muted \citep{xu2016msr}, we have excluded the ASR-based version. ``Event'' refers to the event decomposition (\cref{sec:query_decom}) and ``ASR'' refers to the audio decomposition (\cref{sec:asr}). Results from our whole method is {\hlc[tablegray]{highlighted}}.}
\label{tab:main_result}
\end{table*}

\paragraph{Query vs. Multimodal Descriptions}

In this approach, we used the original query but replaced the video content with the Multimodal Description Set, yielding a text-to-text similarity score instead of text-to-video. Specifically, we compute the maximum similarity between the query and all multimodal descriptions. Given $C$ descriptions associated with the video:
\begin{equation}
    S_{i,v} | Q = \max_{c=1:C} Sim_t(Q, Cap(v)_c), \label{eq:query_vs_caps}
\end{equation}
where $Cap(v)$ returns the multimodal description of the video $v$ and $Sim_t(x, y)$ computes the ColBERT\footnote{ %
We experimented with cosine-based SBERT~\citep{reimers-gurevych-2019-sentence} and found that ColBERT outperforms SBERT in both accuracy and speed. Notably, the captions generated by LLMs may have some noise. ColBERT mitigates the noise's impact by leveraging maximum aggregation at the token level, rather than compressing information into a single vector, which effectively reduces or ignores noisy elements in the multimodal descriptions. Additionally, prior studies \citep{warner2024smarter} have shown that ColBERT’s approach minimizes information loss introduced by the averaging process used in SBERT, further enhancing its performance. %
} similarity score between $x$ and $y$.

\begin{table*}[!ht]
\resizebox{\textwidth}{!}{
    \begin{tabular}{@{}lccccccccccc@{}}
    \toprule
    Model & R@1$\uparrow$ & R@5$\uparrow$ & R@10$\uparrow$ & P@1$\uparrow$ & P@5$\uparrow$ & P@10$\uparrow$ & MRR$\uparrow$ & NDCG$\uparrow$ & MAP$\uparrow$ & MnR$\downarrow$ & MdR$\downarrow$ \\
    \midrule
    & \multicolumn{11}{c}{\textbf{Arabic}} \\
    \cmidrule(lr){2-12}
    MultiCLIP & \textbf{10.91} & 44.32 & 72.31 & \textbf{92.16} & 76.47 & 62.55 & \textbf{0.94} & 76.10 & 84.59 & 14.44 & \textbf{6.0} \\
    \qquad \ablateplus{Event}   & 10.82 & 45.86 & 74.46 & 90.20 & 78.43 & 63.92 & \textbf{0.94} & 78.09 & 86.08 & 14.91 & \textbf{6.0} \\
    \qquad \ablateplus{ASR}  & 10.79 & \textbf{49.66} & \textbf{79.41} & 90.20 & \textbf{83.53} & \textbf{68.24} & \textbf{0.94} & \textbf{82.07} & \textbf{88.32} & \textbf{11.59} & \textbf{6.0} \\
    
    \addlinespace[1mm]

    & \multicolumn{11}{c}{\textbf{Chinese}} \\
    \cmidrule(lr){2-12}
    MultiCLIP & 9.12 & 45.59 & 74.22 & 84.62 & 83.85 & 68.65 & 0.91 & 77.29 & 86.52 & 12.60 & 6.0 \\
    \qquad \ablateplus{Event}   & \textbf{10.66} & 48.81 & 82.13 & \textbf{98.08} & 90.00 & 76.35 & \textbf{0.99} & 86.32 & \textbf{94.64} & 9.59 & 6.0 \\
    \qquad \ablateplus{ASR}  & 10.46 & \textbf{50.55} & \textbf{83.02} & 96.15 & \textbf{93.08} & \textbf{76.92} & 0.97 & \textbf{86.61} & 93.96 & \textbf{8.73} & \textbf{5.5} \\
    
    \addlinespace[1mm]

    & \multicolumn{11}{c}{\textbf{English}} \\
    \cmidrule(lr){2-12}
    MultiCLIP & \textbf{10.61} & 50.82 & 85.79 & \textbf{100.00} & 96.15 & 81.92 & \textbf{1.00} & 89.98 & 97.47 & \textbf{7.64} & \textbf{5.0} \\
    \qquad \ablateplus{Event}   & \textbf{10.61} & \textbf{51.66} & 86.45 & \textbf{100.00} & \textbf{97.69} & 82.50 & \textbf{1.00} & 90.57 & 98.04 & 8.07 & \textbf{5.0} \\
    \qquad \ablateplus{ASR}  & \textbf{10.61} & \textbf{51.66} & \textbf{87.80} & \textbf{100.00} & \textbf{97.69} & \textbf{83.85} & \textbf{1.00} & \textbf{91.43} & \textbf{98.12} & 7.74 & \textbf{5.0} \\
    
    \addlinespace[1mm]

    & \multicolumn{11}{c}{\textbf{Korean}} \\
    \cmidrule(lr){2-12}
    MultiCLIP & 9.30 & 39.98 & 65.33 & 86.54 & 75.00 & 62.12 & 0.89 & 70.38 & 82.86 & 20.78 & 7.0 \\
    \qquad \ablateplus{Event}   & \textbf{9.68} & 43.65 & 72.65 & \textbf{90.38} & 82.31 & 69.04 & \textbf{0.93} & 76.92 & 87.04 & 15.87 & \textbf{6.0} \\
    \qquad \ablateplus{ASR}  & \textbf{9.68} & \textbf{45.93} & \textbf{76.90} & \textbf{90.38} & \textbf{86.15} & \textbf{73.27} & 0.92 & \textbf{80.58} & \textbf{89.02} & \textbf{13.85} & \textbf{6.0} \\
    
    \addlinespace[1mm]

    & \multicolumn{11}{c}{\textbf{Russian}} \\
    \cmidrule(lr){2-12}
    MultiCLIP & 10.57 & 49.57 & 79.09 & 92.31 & 87.69 & 71.15 & 0.96 & 82.84 & 91.20 & 10.65 & 6.0 \\
    \qquad \ablateplus{Event}   & 10.39 & 50.55 & 81.61 & 92.31 & 89.62 & 73.46 & 0.96 & 84.62 & 91.46 & 10.88 & 6.0 \\
    \qquad \ablateplus{ASR}  & \textbf{11.21} & \textbf{53.49} & \textbf{85.40} & \textbf{98.08} & \textbf{94.23} & \textbf{76.54} & \textbf{0.99} & \textbf{88.70} & \textbf{95.02} & \textbf{9.13} & \textbf{5.0} \\
    
    \bottomrule
    \end{tabular}
}

\caption{Comprehensive comparison of different languages. Results are reported on the MultiVENT dataset.}
\label{tab:lg_diff_lang2}
\end{table*}

\paragraph{Event Decomposition vs. Multimodal Descriptions}
We use the refined prequel/current/sequel events from the query (\cref{sec:decom_refinement}) and the multimodal description set (\cref{sec:video_decom}) to compute a many-to-many similarity score. %
Specifically, we compute the maximum similarity score across all descriptions for each event and subsequently take the maximum similarity score among all the events.
Formally:
\begin{equation}
    S_{i,v} | Q = 
    \max_{l=1:L} \max_{c=1:C} Sim_t(E_i(Q)_l, Cap(v)_c),
    \label{eq:events_vs_caps}
\end{equation}
where $L$ is the number of events (prequel / current / sequel) in the query, and $E_i()$ returns the respective event decomposition (prequel / current / sequel).

This global maximum based approach reduces the negative impact of a single noisy decomposition by leveraging the global maximum instead of averaging, which is more vulnerable to noisy data. %
This global maximum effectively decrease the effect of plausible hallucination that can be expected with the LLM generation.
For instance, for the query ``LA Fire'' one of the decomposed prequels might be ``Man is playing with a dog'' (a hallucinated event). As our method generates multiple decomposition of the same event but also takes the global maximum, \ours effectively reduce the effect of hallucinated decompositions.

\paragraph{Score Fusion}
A traditional approach of combining ranks from different experts is to train a fusion model. %
But as \ours is based on zero-shot use of LLM agents, fusing all the scores presents a challenge. Naively averaging all scores would assume equal importance for each score, which does not reflect real-world scenarios. %
E.g., for videos relating to earthquakes, the Sequel vs. Descriptions score carries significantly more weight than the Prequel vs. Descriptions, as there is no visual indication of an earthquake before it occurs. %
\citet{yin2024multi} demonstrated that entropy-based methods outperform others in zero-shot rank fusion tasks. In this entropy ranking, disparate similarity scores for an individual video are normalized to a softmax distribution, where low entropy indicates high confidence, while high inverse entropy corresponds to high confidence. Building on this insight, we use inverse entropy rank fusion to combine the scores. %

Specifically, we convert each of the five previously computed similarity scores $S_{i}$ to a weighted distribution $P_i=\mathrm{softmax}(S_i)$, and compute the entropy $H(P_i)$ of this distribution. This $P_i$ provides an initial likelihood weighting across the $V$ videos, according to the $i$th scoring criteria. We rescale $P_i$ by its inverse entropy, computing a final score $\hat{S}$ by summing across all rescaled $P_i$: %
\begin{equation}
    \hat{S} = \sum_i^5 \frac{1}{H(P_i)} * P_i. \label{eq:rank}
\end{equation}
By applying \cref{eq:rank}, we effectively assign greater weights to scores with lower entropy, emphasizing more confident predictions. %
This $\hat{S}$ contains an aggregate score for each video; $\hat{S}$ can be sorted to return the ranked list of videos for the query.

\section{Results}
\begin{figure*}
    \centering
    \includegraphics[width=0.95\linewidth]{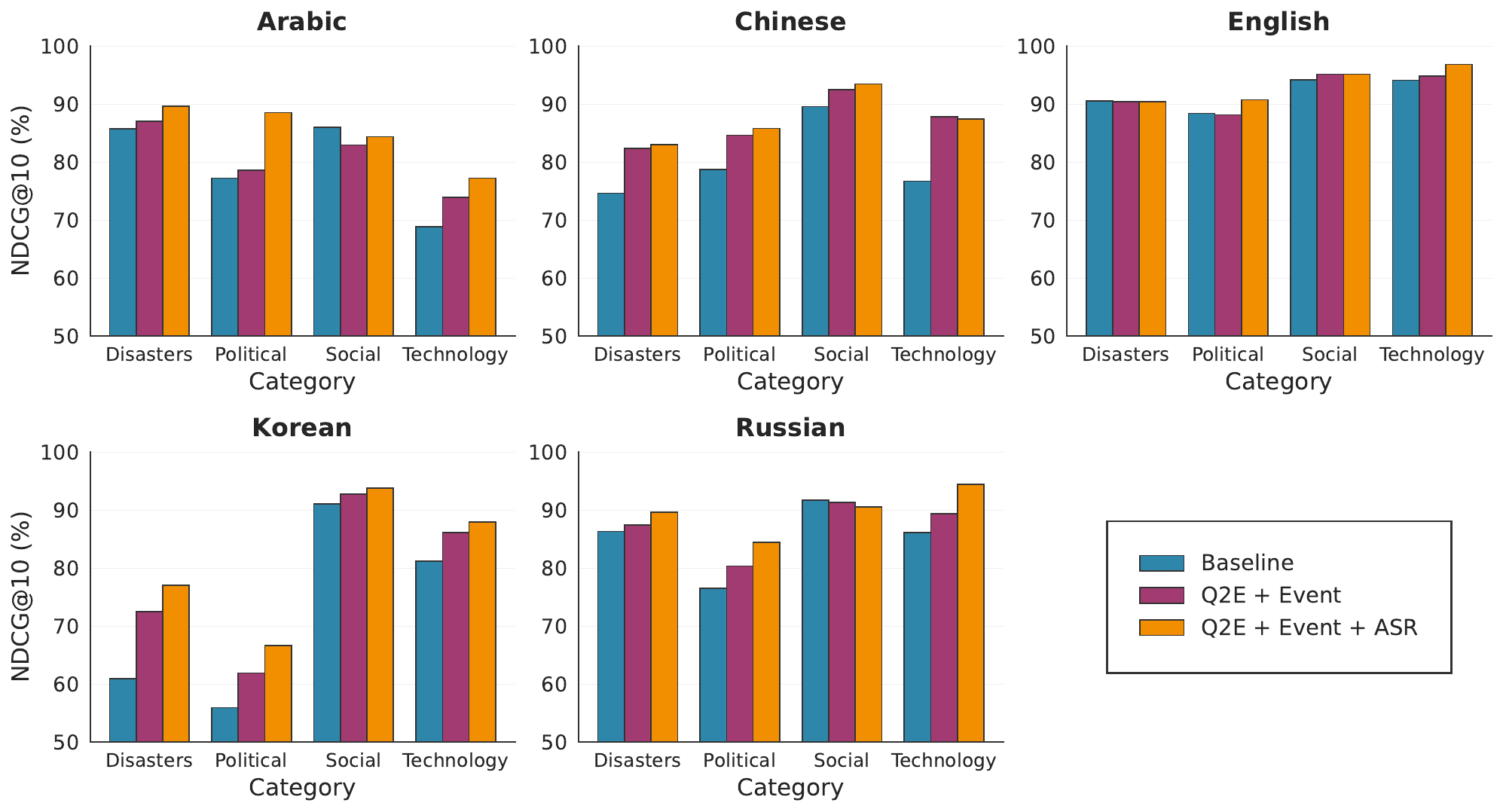}
    \caption{Comparison of NDCG@10 scores across topic categories (Disasters, Political, Social, Technology) for five languages--Arabic, Chinese, English, Korean, and Russian. The results demonstrate the performance improvement from the baseline to our method (with audio or without), highlighting consistent gains across languages and domains.}
    \label{fig:diff_lang_diff_cat}
\vspace{-2mm}
\end{figure*}

\subsection{Datasets and Implementation}
We used the event-centric MultiVENT dataset \citep{sanders2023multivent}, the widely-used MSR-VTT \citep{xu2016msr} and MSVD \citep{chen2011building} dataset. %
We used the standard ``MSR-VTT-1k-A'' test split of 1,000 videos for computational constraint, as used in prior works \citep{jiang2022tencent, guan2023pidro, jiang2023dual}.

Most of the MultiVENT videos are event-specific human queries with complex visual content (i.e., flood, people running) and rapid scene changes (i.e., news reports where multiple scenes are shown), where the average video  is approximately 5.5 times longer than the MSR-VTT dataset (\cref{tab:dataset_specs}). On the other hand, MSR-VTT and MSVD focus on general scenarios of everyday life, i.e., a man is playing with the dog. We have used different datasets to show the robustness of our method across different datasets of complexity. Full implementation details are in \cref{app:implementation}

\subsection{Zero-Shot Text-to-Video Retrieval}
We use MultiCLIP (per \citet{sanders2023multivent}) and InternVideo2 (the best-performing method on the MSR-VTT leaderboard\footnote{\url{https://paperswithcode.com/sota/zero-shot-video-retrieval-on-msr-vtt}}) as our baseline encoders.
The comparison of our method with the baseline for MultiVENT, MSR-VTT-1kA and MSVD is shown on the \cref{tab:main_result}. We use the standard retrieval metrics: 
recall/precision at rank K, mean reciprocal rank (MRR), mean average precision (MAP), normalized discounted cumulative gain (NDCG), and mean and median rank (MnR, MdR).

Overall, {\bfseries combined with event decomposition and ASR, \ours gives the best score in all the metrics across three different datasets and two different text-to-video encoders}. But {\bfseries even without ASR, we outperformed the baseline in all metrics}. %
To show the generalizability of our method across different text-video encoders, we applied it to both the MultiCLIP and InternVideo2 encoder. 

\cref{tab:main_result} shows the generality of our method. When applied to the InternVideo2, our method improved the MSR-VTT NDCG score by 1 point without audio and 4 points with audio. Additionally, the significant performance drop observed on the MultiVENT dataset underscores the dataset’s richness and complexity. Despite this drop, incorporating our method boosted the score by \textbf{19 points} without audio and \textbf{26 points} with audio, further validating its effectiveness.

Conversely, MultiCLIP achieved the best baseline score on the MultiVENT dataset \citep{sanders2023multivent}. Applying our method to MultiCLIP resulted in an \textbf{8-point} NDCG improvement, setting a new state-of-the-art for the MultiVENT dataset. However, unlike InternVideo2, MultiCLIP does not use a cross-encoder reranker, which explains its comparatively lower performance on the MSR-VTT dataset. Our method still led to a \textbf{4-point} improvement in the MSR-VTT NDCG score, demonstrating effectiveness across different encoders. Additionally, a full comparison with recent and popular baselines are provided in \cref{sec:comp_with_baselines}.

Next, we examine \ours's performance across languages (\cref{sec:diff_lang}), the effect of different LLM sizes (\cref{sec:llm_sizes}), different rank fusion methods (\cref{sec:rank_fusion}) and various components of our method (\cref{sec:diff_components}). In the appendix, we examine the impact of VLM sizes (\cref{sec:app_vlm_sizes}), different layers of ASR module (\cref{sec:app_diff_layer_translation}), number of frames (\cref{sec:app_num_of_frames}), and frame selection methods (\cref{sec:app_frame_selection_method}). We provide a qualitative analysis in \cref{app:qual_analysis}.

\subsection{Performance Across Languages} \label{sec:diff_lang}
MultiVENT includes five languages -- Arabic, Chinese, English, Korean, and Russian. \cref{tab:lg_diff_lang2} shows that building upon the MultiCLIP baseline, incorporating \ours\ consistently improves retrieval performance across all metrics. 
outperforming the baseline by high numbers. 
We observed that audio-based decomposition plays a more significant role in languages like Arabic and Russian, whereas event-based decomposition is more crucial for Chinese and Korean. However, when combined, the two approaches complement each other, resulting in an aggregated performance boost.

MultiCLIP struggles the most with Arabic, Chinese, and Korean video retrieval. However, when our method is added, retrieval performance improves by approximately 6, 9, and 10 NDCG points for these languages, respectively. 

To analyze the bias of the events and the effect of the bias on the performance of our method, we have used event category (provided in the main dataset) to divide the queries into 4 different buckets -- Disaster, Political, Social, Technology. The results of different bucket across different language is provided in the \cref{fig:diff_lang_diff_cat}. We can see that our method improves the NDCG score irrespective of language or event category (except Arabic-Social). Also, the performance improvement is boosted with addition of audio features. The Social category, in particular, shows the most variation, with Arabic and Russian even experiencing a slight dip in performance, which could be linked to cultural differences in how queries are interpreted. These varying results across languages support the idea that query decomposition brings out language-specific nuances.

\begin{table}[t!]
\resizebox{\columnwidth}{!}{
    \begin{tabular}{@{}lcccc@{}}
    \toprule
    Size & R@10$\uparrow$ & P@10$\uparrow$ & MRR$\uparrow$ & NDCG$\uparrow$ \\ 
    \midrule
    MultiCLIP & 70.82 & 65.25 & 0.92 & 75.34 \\

    \cmidrule(lr){2-5}
    1B & 78.71 & 72.28 & \textbf{0.95} & 82.50 \\
    3B & 79.17 & 72.74 & \textbf{0.95} & 83.03 \\
    8B & 79.04 & 72.59 & \textbf{0.95} & 82.91 \\
    70B (\ours) & \textbf{79.60} & \textbf{73.09} & \textbf{0.95} & \textbf{83.24} \\
    \bottomrule
    \end{tabular}
}

\caption{Comparison of different LLM sizes within the Llama-3 family (1B to 70B parameters), specifically used for query decomposition and refining frame captions to video caption. Results are reported on the MultiVENT dataset. %
Full results in \cref{tab:lg_llm_sizes}, \cref{sec:app_llm_sizes}.
}
\label{tab:sm_llm_sizes}
\vspace{-2mm}
\end{table}

\subsection{Effect of LLM size} \label{sec:llm_sizes}
Our original method used the \texttt{Llama-3.3-70B} LLM. While this model is highly robust and generalizable, its large size demands significant computational power and increased inference time. To demonstrate the robustness of our method, \cref{tab:sm_llm_sizes} presents results for different LLM sizes, ranging from 1B to 70B parameters. %
Although the best performance is typically achieved with the 70B model, even the 1B model consistently outperforms the baseline NDCG by at least 8 points. These results demonstrate that our method can achieve comparable performance even when using significantly smaller LLMs, highlighting its practicality for real-world retrieval systems.

\subsection{Effect of Rank Fusion Approaches} \label{sec:rank_fusion}
\begin{table}[t!]
\resizebox{\columnwidth}{!}{
    \begin{tabular}{@{}lcccc@{}}
    \toprule
    Method & R@10$\uparrow$ & P@10$\uparrow$ & MRR$\uparrow$ & NDCG$\uparrow$ \\
    \midrule

    Neg. Exp. Ent. & 67.23 & 61.97 & 0.93 & 73.20 \\
    RRF & 70.91 & 65.29 & 0.93 & 76.29 \\
    Max & 76.10 & 70.04 & 0.93 & 80.04 \\
    Mean & 78.64 & 72.47 & \textbf{0.95} & 82.44 \\
    Inv. Ent. (\ours) & \textbf{79.60} & \textbf{73.09} & \textbf{0.95} & \textbf{83.24} \\
    \bottomrule
    \end{tabular}
}

\caption{Comparison of Zero-Shot Rank Fusion methods: Neg. Exp. Ent. (Negative Exponential Entropy), RRF (Reciprocal Rank Fusion), Inv. Ent. (Inverse Entropy), Mean and Max. Results are reported on the MultiVENT dataset. Full results in \cref{tab:lg_agg_methods}, \cref{sec:app_rank_fusion}.}
\label{tab:sm_agg_methods}
\vspace{-4mm}
\end{table}

Fusing different ranks in a zero-shot setting is a non-trivial task. Traditional approaches often involve training dense networks to combine scores \citep{yu-etal-2024-fusion} or leveraging LLMs to ensemble multiple expert outputs \citep{lu-etal-2024-experts, lu-etal-2024-routing, jiang2023llm}. However, rank or score fusion is a significant challenge for LLMs, as they must process and compare a large volume of numerical data, including floating-point values (for scores) and discrete values (for ranks). In both cases, LLMs struggle to produce reasonable fusion ranks. %
We considered several aggregation methods, from simple techniques like Mean to more complex approaches such as Reciprocal Rank Fusion (RRF) and Inverse Entropy; due to space, we describe these formulations in \cref{sec:app_rank_fusion}. Results in \cref{tab:sm_agg_methods} indicate that even the simple Mean function performs well compared to RRF and Negative Exponential Entropy. Inverse Entropy demonstrated better performance, leading us to adopt it as the aggregation method.

While we adopted global max aggregation, for a fine-grained analysis of the max and mean operator, we have also tried different combination of max-mean aggregation over captions and events. The results are shown in the \cref{tab:aggregation_results}. The results show that (1) \ours performs the best compared with other aggregation methods.
(2) Mean over captions underperformed the most, specifically learning nothing. We hypothesize that, as previously seen, the VLM models make noisy captions, and averaging over them hurts the model beyond what can be overcome.
(3) Most of the other methods perform comparably with the global max, though we think global max wins because of its small performance gain and high simplicity.
(4) The consistent performance across different evaluation metrics indicates that our method offers reliable improvements rather than metric-specific optimizations.

\subsection{Effect of Different Components} \label{sec:diff_components}

\begin{table}[t]
\resizebox{\columnwidth}{!}{
    \begin{tabular}{@{}lcccc@{}}
    \toprule
    Components & R@10$\uparrow$ & P@10$\uparrow$ & MRR$\uparrow$ & NDCG$\uparrow$ \\
    \midrule

    \ours & \textbf{79.60} & \textbf{73.09} & \textbf{0.95} & \textbf{83.24} \\
    \quad \ablate{Video} & 67.74 & 62.43 & 0.93 & 73.96 \\
    \quad \ablate{Query} & 78.12 & 71.74 & 0.93 & 81.54 \\
    \quad \ablate{Event} & 77.77 & 71.47 & 0.94 & 81.75 \\
    \bottomrule
    \end{tabular}
}

\caption{Analysis of different components in our method. %
Event = (Prequel + Current + Sequel). Results are reported on the MultiVENT dataset. The full result is reported on \cref{tab:lg_diff_components} in \cref{sec:app_diff_components}.}
\label{tab:sm_diff_components}
\vspace{-4mm}
\end{table}

\cref{tab:sm_diff_components} presents an ablation study of event decomposition and multimodal description components of \ours. As expected, the results show that the video is the most critical.
Removing the query results in a 1-point performance drop, while removing events reduces performance by 2 points, highlighting the importance of events for retrieving relevant videos.

\section{Related Work}

With the rapid advancements in LLMs~\citep{devlin2019bert,radford2019language,touvron2023llama,dubey2024llama,gunasekar2023textbooks,liu2024deepseek,jiang2024mixtral,chu2024qwen2}, methodologies such as zero-shot, few-shot, and in-context learning have gained substantial traction due to their reduced computational requirements and enhanced time efficiency. %
These have influenced approaches for event causal reasoning, multimodal language models, and retrieval.

\subsection{Event Causal Reasoning}
Reasoning based on event causality plays a crucial role across various domains. \citet{sun-etal-2024-event} demonstrated that event-causal graphs, even generated by LLMs, can enhance story understanding and better align with human evaluations. Similarly, in \ours, we leverage causation from human queries to construct prequel, current, and sequel events, enabling a deeper understanding of user intent. While \citet{chan-etal-2024-exploring} utilized LLMs to infer temporal, causal, and discourse relationships between two events, our approach applies the same temporal (prequel or sequel) and causal (current) reasoning to extract relevant events directly from queries.

\subsection{Vision Language Model}
Vision-Language Models (VLMs) have been used significantly in text-to-video retrieval tasks by leveraging large-scale pre-trained models trained on image-text pairs \citep{ibrahimi2023audio}. These models have demonstrated impressive zero-shot capabilities and transfer learning potential for video-related tasks \citep{zhao2024distilling}. Recent advancements have focused on efficiently adapting image-based VLMs to the video domain, addressing temporal dynamics and computational costs \citep{nishimuravision}. Some approaches propose novel architectures like sparse-and-correlated adapters \citep{cao2024rap} to enhance retrieval performance while maintaining efficiency. Additionally, researchers have explored incorporating audio information \citep{ibrahimi2023audio} and using synthesized instructional data \citep{zhao2024distilling} to improve video-language models further. %

\subsection{Zero-Shot Text-to-Video Retrieval}
Though much has been done in the domain of fine-tuned text-to-video retrieval \citep{wang2025internvideo2, cicchetti2024gramian, chen_vast_2023, xu2023mplug}, there is a gap in zero-shot text-to-video retrieval performance. Fine-tuning a model on a task-specific dataset can lead to impressive performance, but it is not always suitable due to high computing costs and time constraints. While zero-shot text-to-video retrieval has been previously studied~\citep{wang2025internvideo2, cicchetti2024gramian, chen_vast_2023, xu2023mplug} results suggest a sizable gap between that and fine-tuned retrieval.

\section{Conclusion \& Future Work}
We leveraged the world knowledge embedded in LLMs and VLMs to decompose both queries and videos for the video retrieval task, demonstrating how to match this knowledge against visual and audio signals. We employed inverse entropy as a rank fusion method without fine-tuning any additional fusion network. Experimental results across three datasets with different characteristics demonstrate the robustness and generalizability of our approach. 
While we show how the preexisting knowledge of LLMs and VLMs can be effectively transferred to text-to-video retrieval, our work sets the stage for future research to explore how factual and counterfactual information can be leveraged to positively or negatively align with the query. %

\section{Limitations}
While our approach demonstrates superior performance across multiple datasets, it is computationally expensive and time-intensive. Future research should focus on optimizing efficiency to reduce both time and cost.

Due to the LLM and VLM's inherent nature, some amount of misinformation or hallucination may be possible. Although we did not find any misinformation during our evaluation, our prompting of what \textit{could} happen before, during, or after an event could result in hallucinations. Similarly, by relying on the parametric knowledge of these models, there is a potential for biases inherent in the models or their outputs to be propagated. We did not explicitly explore or mitigate the biases inherent in these models, though we also did not notice objectionable or stereotypical output.

The primary focus of our method is to demonstrate how query enrichment through decomposition can improve understanding of human intent and ground specific events to video content. As we have not applied any enhancements to the video captions or audio transcripts, video-to-text retrieval is outside the scope of this work.

\section{Acknowledgment}
This material is based in part upon work supported by the National Science Foundation under Grant No. IIS-2024878. %
Some experiments were conducted on the UMBC HPCF, supported by the National Science Foundation under Grant No. CNS-1920079. %
This material is also based on research that is in part supported by the Army Research Laboratory, Grant No. W911NF2120076, and by DARPA for the SciFy program under agreement number HR00112520301. The U.S. Government is authorized to reproduce and distribute reprints for Governmental purposes notwithstanding any copyright notation thereon. The views and conclusions contained herein are those of the authors and should not be interpreted as necessarily representing the official policies or endorsements, either express or implied, of ARL, DARPA or the U.S. Government.

\bibliography{anthology,custom,dipta}

\appendix
\section{Appendix} \label{sec:appendix}

\subsection{Comparison with Popular Baselines} \label{sec:comp_with_baselines}

To show the robustness of our zero-shot method, in the \cref{tab:comparison_with_baselins}, we have reported the comparison of our method with different recent models: Frozen in Time \citep{bain_frozen_2022}, ImageBind \citep{girdhar_imagebind_2023}, CLIP \citep{radford2021learning}, CLIP4CLIP \citep{luo_clip4clip_2021}, CLIP2Video \citep{fang_clip2video_2021}, X-Pool \citep{gorti_x-pool_2022}, TS2-Net \citep{liu_ts2-net_2022}, TDB+TAB \citep{fang_transferring_2023}, IVRC \citep{tian_rethink_2024}, Cap4Video \citep{wu_cap4video_2023}, DRL \citep{wang_disentangled_2022}, T2VIndexer \citep{li_t2vindexer_2024}, PIDRo \citep{guan_pidro_2023}, CLIP-VIP \citep{xue_clip-vip_2023}, huge \citep{jiang2022tencent}.

\begin{table}[!ht]
\centering
\resizebox{\columnwidth}{!}{%
    \begin{tabular}{@{}lccccr@{}}
        \toprule
        Models & Venue & FT & R@1$\uparrow$ & R@5$\uparrow$ & R@10$\uparrow$ \\
        \midrule
        & \multicolumn{5}{c}{\textbf{MSR-VTT}} \\
        \cmidrule(lr){2-6}
        Frozen in Time & ICCV '21 & $\times$ & 27.30 & - & 68.10 \\
        ImageBind & CVPR '23 & $\times$ & 36.80 & 61.80 & 70.00 \\
        CLIP & PMLR '21 & $\times$ & 39.70 & 72.30 & 82.20 \\
        CLIP4CLIP & arXiv '21 & $\checkmark$ & 44.50 & 71.40 & 81.60 \\
        CLIP2Video & arXiv '21 & $\checkmark$ & 45.60 & 72.60 & 81.70 \\
        X-Pool & CVPR '22 & $\checkmark$ & 46.90 & 72.80 & 82.20 \\
        TS2-Net & ECCV '22 & $\checkmark$ & 45.60 & 74.40 & 82.70 \\
        TDB+TAB & TMM '23 & $\checkmark$ & 45.60 & 72.60 & 81.70 \\
        IVRC & Patt. Recog. '24 & $\checkmark$ & 47.00 & 75.00 & 82.40 \\
        Cap4Video & CVPR '23 & $\checkmark$ & 51.40 & 75.70 & 83.90 \\
        DRL & arXiv '22 & $\checkmark$ & 53.30 & 80.30 & 87.60 \\
        T2VIndexer & ACM MM '24 & $\checkmark$ & 55.10 & 77.20 & 85.00 \\
        PIDRo & ICCV '23 & $\checkmark$ & 55.90 & 79.80 & 87.60 \\
        \rowcolor{tablegray} \ours & - & $\times$ & 56.28 & 76.58 & 83.72 \\
        CLIP-ViP & ICLR '23 & $\checkmark$ & 57.70 & 80.50 & 88.20 \\
        huge & arXiv '22 & $\checkmark$ & \textbf{62.90} & \textbf{84.50} & \textbf{90.80} \\
        
        & \multicolumn{5}{c}{\textbf{MSVD}} \\
        \cmidrule(lr){2-6}
        Frozen in Time & ICCV '21 & $\times$ & 33.70 & 64.70 & 76.30 \\
        CLIP & PMLR '21 & $\times$ & 37.00 & 64.10 & 73.80 \\
        CLIP4CLIP & arXiv '21 & $\checkmark$ & 46.20 & 76.10 & 84.60 \\
        TDB+TAB & TMM '23 & $\checkmark$ & 46.90 & 76.90 & 86.10 \\
        CLIP2Video & arXiv '21 & $\checkmark$ & 47.00 & 76.80 & 85.90 \\
        X-Pool & CVPR '22 & $\checkmark$ & 47.20 & 77.40 & 86.00 \\
        PIDRo & ICCV '23 & $\checkmark$ & 47.50 & 77.50 & 86.00 \\
        DRL & arXiv '22 & $\checkmark$ & 48.30 & 79.10 & 87.30 \\
        Cap4Video & CVPR '23 & $\checkmark$ & 51.80 & 80.80 & 88.30 \\
        T2VIndexer & ACM MM '24 & $\checkmark$ & 55.40 & 81.90 & 88.50 \\
        huge & arXiv '22 & $\checkmark$ & 59.00 & 84.00 & \textbf{90.30} \\
        \rowcolor{tablegray} \ours & - & $\times$ & \textbf{63.77} & \textbf{85.30} & 89.99 \\
        \bottomrule
    \end{tabular}
}
\caption{Results from different baseline models reported in the recent papers. Rows are sorted based on R@1 score. FT represents the fine-tuned models for text-to-video retrieval task.}
\label{tab:comparison_with_baselins}
\end{table}

On the \textbf{MSR-VTT} dataset, our method outperforms all of the zero-shot methods by high margin. However, while fine-tuning can achieve moderately higher performance, our zero-shot method performs competitively. 
On the other hand, in the \textbf{MSVD} dataset, our method outperforms all baseline models, whether fine-tuned or not. This again underlines the generalizability and robustness of our method across different datasets compared to the previous baselines.

\subsection{Qualitative Analysis} \label{app:qual_analysis}
We have reported two examples from MSRVTT \citep{xu2016msr} and MultiVENT \citep{sanders2023multivent} dataset on the \cref{tab:qual_analysis}. %
We observed that while event decomposition may not be as grounded or clearly delineated as in the MultiVENT dataset (due to the inherent ungrounded nature of MSRVTT queries), our approach still enriches the query by introducing events that could occur before, during or after the main event. This enrichment significantly enhances the retrieval process. 

\begin{table*}[!ht]
    \centering
    \small
    \resizebox{0.95\textwidth}{!}{%
    \begin{tabular}{p{0.1\textwidth}|p{0.90\textwidth}}
    \toprule
    \rowcolor{tablegray} \textbf{Dataset} & \textbf{MSR-VTT} \\
    
    \vspace{0.1mm} \\
    Query & Crocodiles about to eat the group of people walking across the water \\

    \vspace{0.1mm} \\
    Prequel &
\vspace{-4.7mm}
\begin{itemize}[leftmargin=*]
\setlength\itemsep{0pt}
\item People wading through the water, unaware of the crocodiles before they attack
\item Crocodiles swimming stealthily towards the group during an attack on people
\item The group noticing the crocodiles and trying to hurry across the water 
\item Crocodiles rising out of the water, jaws open 
\item People walking cautiously across the water to avoid crocodiles attacking them
\end{itemize} \\

    Current &
\vspace{-4.7mm}
\begin{itemize}[leftmargin=*]
\setlength\itemsep{0pt}
\item Crocodiles attacking people while swimming towards a group of them
\item Crocodiles emerging from the water with their mouths open during attacks on people
\item People screaming and trying to run or swim away from crocodiles during attacks on humans
\item People trying to hurry or run across the water when crocodiles attack them
\item Crocodiles snapping their jaws near people during attacks
\end{itemize} \\
    
    Sequel &
\vspace{-4.7mm}
\begin{itemize}[leftmargin=*]
\setlength\itemsep{0pt}
\item Crocodiles attacking people while chasing after a group
\item People defending themselves against crocodile attacks when crocodiles attack them
\item Crocodiles dragging the people underwater
\end{itemize} \\

    \midrule
    \midrule
    \rowcolor{tablegray} \textbf{Dataset} & \textbf{MultiVENT} \\
    
    \vspace{0.1mm} \\
    Query & November 30 earthquake in South Central Alaska 2018 \\
    
        \vspace{0.1mm} \\
        Prequel &
\vspace{-4.7mm}
\begin{itemize}[leftmargin=*]
\setlength\itemsep{0pt}
\item Buildings shaking and swaying due to the earthquake in Anchorage, Alaska, on November 30, 2018
\item People running out of buildings and evacuating the area in Anchorage, Alaska during the earthquake on November 30, 2018
\item Cars stopped on the road as the earthquake strikes in Anchorage, Alaska on November 30, 2018
\item Debris and objects falling from shelves and ceilings during the 30 November 2018 Anchorage, Alaska earthquake
\item Emergency responders rushing to the scene to assist with evacuation and relief efforts after the 30 November 2018 earthquake in Anchorage, Alaska, South Central Alaska
\end{itemize} \\
    
        Current &
\vspace{-4.7mm}
\begin{itemize}[leftmargin=*]
\setlength\itemsep{0pt}
\item Buildings shaking and crumbling during the 30 November 2018 earthquake in Anchorage, Alaska, South Central Alaska
\item People running out of buildings and evacuating the area during the 30 November 2018 Anchorage, Alaska earthquake
\item Emergency responders rushing to the scene after the 2018 Anchorage, Alaska earthquake on November 30, 2018
\item Debris falling from buildings and damaging streets during the November 30, 2018 earthquake in Anchorage, South Central Alaska
\item Cars stopped or abandoned on the road in Anchorage, Alaska, South Central Alaska due to the earthquake on November 30, 2018
\end{itemize} \\
        
        Sequel &
\vspace{-4.7mm}
\begin{itemize}[leftmargin=*]
\setlength\itemsep{0pt}
\item Buildings crumbling or collapsing during the November 30, 2018 earthquake in Anchorage, Alaska
\item People running for cover or evacuating buildings during the November 30, 2018 earthquake in Anchorage, Alaska
\item Emergency vehicles rushing to the scene of the 2018 Anchorage, Alaska earthquake on November 30, 2018
\item Cracks forming in roads and highways in Anchorage, Alaska, South Central Alaska, after the 30 November 2018 earthquake
\item Debris falling from damaged structures during the 30 November 2018 earthquake in Anchorage, Alaska
\end{itemize} \\

    \bottomrule
    \end{tabular}
    }
    \caption{Qualitative analysis on two examples from MSRVTT \citep{xu2016msr} and MultiVENT\citep{sanders2023multivent} dataset respectively.}
    \label{tab:qual_analysis}
\end{table*}

From the retrieval ranking perspective, incorporating event-based decomposition substantially improves video retrieval accuracy. For example, for the provided MSR-VTT query, the ground-truth video is ranked 12th (with audio) and 13th (without audio) by MultiCLIP, whereas \ours (which integrates decomposition module, multimodal features, textual descriptions, and our ASR module) elevates it to 1st place with audio and 7th place without audio.

\subsection{Ablation Study (continued)} \label{sec:ablation}

\subsubsection{Effect of VLM Sizes} \label{sec:app_vlm_sizes}
\begin{table*}[!ht]
\resizebox{\textwidth}{!}{
    \begin{tabular}{@{}lccccccccccc@{}}
    \toprule
    VLM Size & R@1$\uparrow$ & R@5$\uparrow$ & R@10$\uparrow$ & P@1$\uparrow$ & P@5$\uparrow$ & P@10$\uparrow$ & MRR$\uparrow$ & NDCG$\uparrow$ & MAP$\uparrow$ & MnR$\downarrow$ & MdR$\downarrow$ \\
    \midrule
    MultiCLIP & 9.83 & 44.32 & 70.82 & 88.80 & 80.77 & 65.25 & 0.92 & 75.34 & 86.33 & 22.12 & \textbf{6} \\
    \addlinespace[1mm]
    
    & \multicolumn{11}{c}{\textbf{Without Audio}} \\
    \cmidrule(lr){2-12}
    1B & 9.80 & 43.76 & 71.73 & 89.19 & 79.85 & 66.02 & 0.93 & 75.86 & 86.05 & 24.96 & \textbf{6} \\
    2B & 9.94 & 45.23 & 73.86 & 89.96 & 82.55 & 68.03 & 0.93 & 77.90 & 87.49 & 22.45 & \textbf{6} \\
    4B & 9.97 & 46.75 & 75.85 & 90.73 & 85.17 & 69.92 & 0.93 & 79.81 & 89.05 & \textbf{19.68} & \textbf{6} \\
    8B & 10.03 & 46.40 & 75.73 & 90.35 & 84.71 & 69.81 & 0.93 & 79.72 & 88.98 & 21.24 & \textbf{6} \\
    26B & \textbf{10.24} & \textbf{47.03} & \textbf{76.16} & \textbf{92.66} & \textbf{85.64} & \textbf{70.19} & \textbf{0.95} & \textbf{80.47} & \textbf{89.94} & 19.87 & \textbf{6} \\
    38B (\ours) & \textbf{10.24} & 46.71 & 75.76 & 92.28 & 85.25 & 69.73 & \textbf{0.95} & 80.04 & 89.42 & 21.13 & \textbf{6} \\
    \addlinespace[1mm]
    
    & \multicolumn{11}{c}{\textbf{With Audio}} \\
    \cmidrule(lr){2-12}
    1B & 10.18 & 47.87 & 78.15 & 91.89 & 86.56 & 71.74 & 0.94 & 81.77 & 90.33 & 18.79 & \textbf{6} \\
    2B & 10.14 & 48.27 & 78.46 & 91.51 & 87.26 & 72.05 & 0.94 & 82.14 & 90.58 & 17.21 & \textbf{6} \\
    4B & 10.19 & 48.90 & 79.72 & 92.28 & 88.57 & 73.17 & 0.94 & 83.13 & 91.05 & 16.77 & \textbf{6} \\
    8B & \textbf{10.40} & 48.87 & 79.88 & \textbf{93.82} & 88.34 & 73.36 & \textbf{0.95} & 83.52 & \textbf{91.53} & 17.27 & \textbf{6} \\
    26B & \textbf{10.40} & 48.64 & \textbf{80.12} & \textbf{93.82} & 88.03 & \textbf{73.51} & \textbf{0.95} & \textbf{83.61} & 91.13 & \textbf{15.97} & \textbf{6} \\
    38B (\ours) & 10.32 & \textbf{49.00} & 79.60 & 93.05 & \textbf{88.73} & 73.09 & \textbf{0.95} & 83.24 & 91.20 & 16.44 & \textbf{6} \\
    \bottomrule
    \end{tabular}
}

\caption{Comprehensive comparison of different VLM sizes within the InternVL-2.5 family (ranging from 1B to 38B parameters), specifically used for frame captioning. Results are reported on the MultiVENT dataset.}
\label{tab:lg_vlm_sizes}
\end{table*}

In \ours, we utilized the \texttt{InternVL-2.5-78B} model. Similar to the \texttt{Llama-3.3-70B}, this model is highly robust and generalizable but comes with significant computational and time constraints. To assess the efficiency and flexibility of our method, \cref{tab:lg_vlm_sizes} reports results across various VLM sizes within the InternVL-2.5 family.

The results indicate that models ranging from 4B to 38B parameters achieve nearly identical performance, with negligible differences in scores. This demonstrates that our method can achieve comparable results even with smaller VLMs, reducing computational requirements. Furthermore, with the inclusion of audio, the performance gap narrows even further; for instance, the 1B VLM experiences only a 1-point drop in NDCG score. These findings underscore the robustness and generalizability of our method, regardless of the VLM size.

\subsubsection{Effect of different layers in Translation Module} \label{sec:app_diff_layer_translation}
\begin{table}[!ht]
\resizebox{\columnwidth}{!}{
    \begin{tabular}{@{}lcccc@{}}
    \toprule
    Components & R@10$\uparrow$ & P@10$\uparrow$ & MRR$\uparrow$ & NDCG$\uparrow$ \\
    \midrule
    
    \ours & \textbf{79.60} & \textbf{73.09} & \textbf{0.95} & \textbf{83.24} \\
    \quad \ablate{ASR} & 79.30 & 72.86 & \textbf{0.95} & 83.02 \\
    \quad \ablate{Translator} & 79.47 & 73.01 & \textbf{0.95} & 83.14 \\
    \quad \ablate{Refiner} & 79.25 & 72.82 & \textbf{0.95} & 82.95 \\
    \quad \ablate{(ASR + Translator + Refiner)} & 75.76 & 69.73 & \textbf{0.95} & 80.04 \\

    \bottomrule
    \end{tabular}
}

\caption{Analysis of different layers of the translation module. Results are reported on the MultiVENT dataset.}
\label{tab:sm_diff_layer_translation}
\end{table}

As described in the \cref{sec:asr}, in our method we have used a multi-layer ASR system consisting of ASR, an automatic translator and an LLM refiner. %
To assess the impact of different layers of translation, we have provided ablation studies in \cref{tab:sm_diff_layer_translation}. In summary, while excluding one translation layer results in a small reduction in NDCG score (0.1-0.5 points), excluding all translation layers leads to a much more significant decline (2-7 points). This suggests strong complementary aspects and a strong aggregate effect.

\subsubsection{Effect of Number of Frames} \label{sec:app_num_of_frames}

\begin{table*}[!ht]
\resizebox{\textwidth}{!}{
    \begin{tabular}{@{}lccccccccccc@{}}
    \toprule
    \# frames & R@1$\uparrow$ & R@5$\uparrow$ & R@10$\uparrow$ & P@1$\uparrow$ & P@5$\uparrow$ & P@10$\uparrow$ & MRR$\uparrow$ & NDCG$\uparrow$ & MAP$\uparrow$ & MnR$\downarrow$ & MdR$\downarrow$ \\
    \midrule
    MultiCLIP & 9.83 & 44.32 & 70.82 & 88.80 & 80.77 & 65.25 & 0.92 & 75.34 & 86.33 & 22.12 & \textbf{6} \\
    \addlinespace[1mm]
    
    & \multicolumn{11}{c}{\textbf{Without Audio}} \\
    \cmidrule(lr){2-12}
    2 & 9.31 & 41.71 & 64.46 & 83.78 & 76.37 & 59.54 & 0.90 & 69.64 & 83.75 & 36.17 & 7 \\
    4 & 9.75 & 44.00 & 71.05 & 88.03 & 80.39 & 65.64 & 0.92 & 75.43 & 86.54 & 25.32 & \textbf{6} \\
    8 & 10.10 & 46.56 & 74.84 & 90.73 & 84.79 & 68.88 & 0.94 & 79.05 & 89.05 & 23.01 & \textbf{6} \\
    16 & \textbf{10.24} & 46.71 & 75.76 & \textbf{92.28} & 85.25 & 69.73 & \textbf{0.95} & 80.04 & 89.42 & 21.13 & \textbf{6} \\
    32 (\ours) & 10.15 & \textbf{47.22} & 75.95 & 91.89 & \textbf{85.95} & 70.00 & 0.94 & 80.29 & \textbf{90.15} & 20.05 & \textbf{6} \\
    64 & 10.09 & 47.15 & \textbf{76.77} & 91.12 & \textbf{85.95} & \textbf{70.66} & 0.94 & \textbf{80.66} & 89.74 & \textbf{19.98} & \textbf{6} \\
    \addlinespace[1mm]
    
    & \multicolumn{11}{c}{\textbf{With Audio}} \\
    \cmidrule(lr){2-12}
    2 & 10.26 & 46.63 & 74.93 & 92.28 & 84.48 & 68.88 & 0.94 & 79.42 & 89.33 & 21.66 & \textbf{6} \\
    4 & 10.16 & 47.37 & 77.54 & 91.51 & 85.79 & 71.27 & 0.94 & 81.26 & 90.05 & 19.20 & \textbf{6} \\
    8 & 10.26 & 48.74 & 79.39 & 92.66 & 88.11 & 72.86 & \textbf{0.95} & 82.93 & 90.99 & 16.97 & \textbf{6} \\
    16 & 10.32 & \textbf{49.00} & 79.60 & 93.05 & \textbf{88.73} & 73.09 & \textbf{0.95} & 83.24 & 91.20 & 16.44 & \textbf{6} \\
    32 (\ours) & \textbf{10.41} & 48.83 & 80.09 & \textbf{93.82} & 88.26 & 73.55 & \textbf{0.95} & 83.68 & 91.32 & 16.38 & \textbf{6} \\
    64 & 10.40 & 48.90 & \textbf{80.42} & \textbf{93.82} & 88.49 & \textbf{73.82} & \textbf{0.95} & \textbf{83.95} & \textbf{91.58} & \textbf{15.90} & \textbf{6} \\
    \bottomrule
    \end{tabular}
}

\caption{Full comparison of different number of frames (ranging from 2--64). Results are reported on the MultiVENT dataset.}
\label{tab:lg_num_of_frames}
\end{table*}

\begin{figure}[!ht]
    \centering
    \includegraphics[width=\columnwidth]{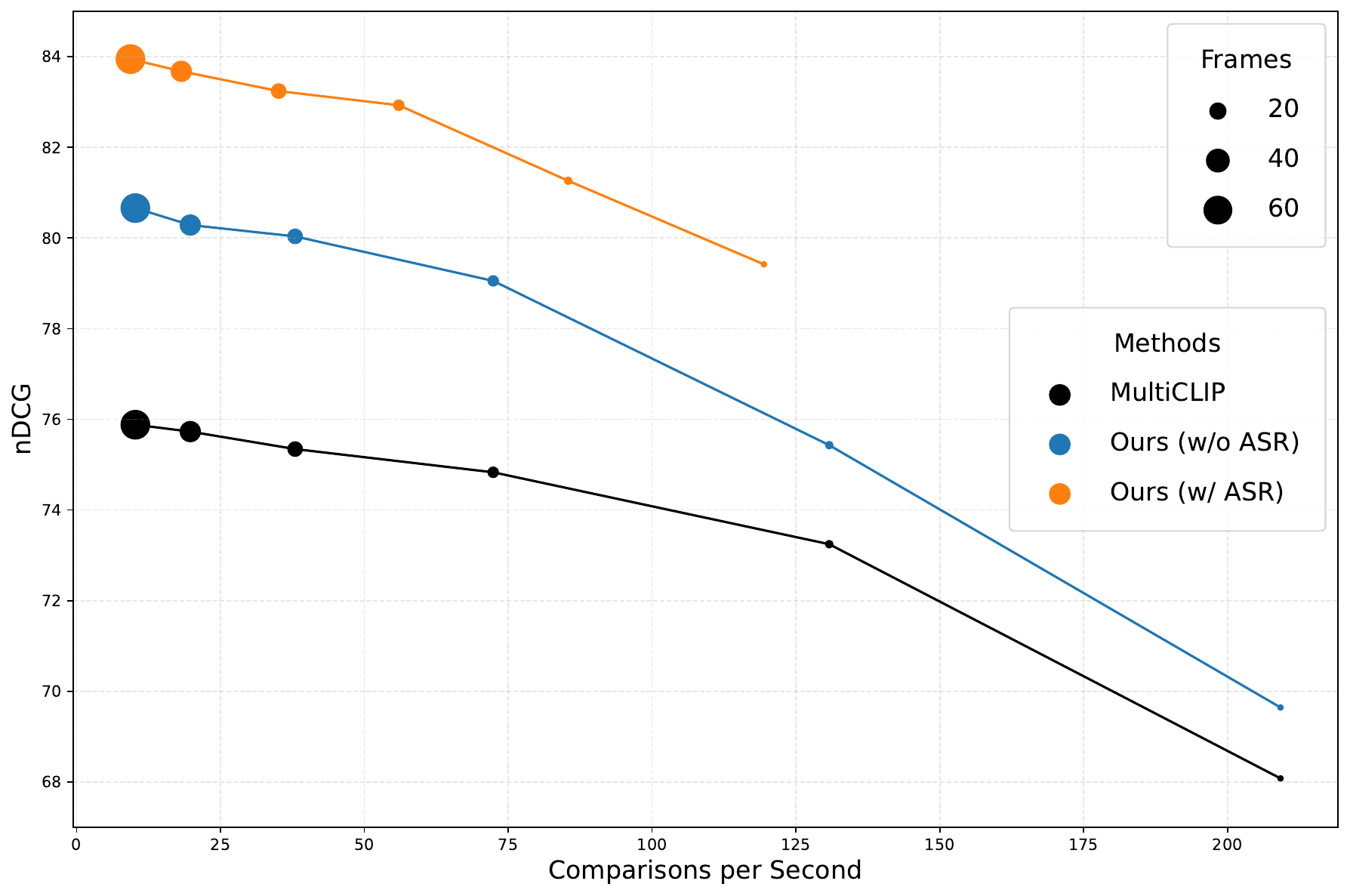}
\caption{ %
Impact of the number of frames on retrieval performance. The figure shows the nDCG performance against comparisons per second for three methods: MultiCLIP (black), \ours\ (without ASR) (blue), and the \ours\ (with ASR) (orange). The point size is proportional to the number of sampled frames (i.e., 2, 4, 8, 16, 32, 64). Comparison per Second = Runtime / (num. of Query $\times$ num. of Video). Results are reported on the MultiVENT dataset.
The full table is reported on \cref{tab:lg_num_of_frames}.
\label{fig:ndcg_vs_time}
}
\end{figure}

\cref{fig:ndcg_vs_time} illustrates the impact of the number of frames on performance across different methods. The results show that increasing the number of sampled frames generally improves performance. However, as the number of comparisons per second increases (i.e., fewer frames are sampled), NDCG declines across all methods, highlighting a trade-off between computational efficiency and retrieval performance.

While the upward trend continues beyond 64 frames, we hypothesize that further increasing the number of frames would yield additional performance gains. However, the diminishing returns and increased computational cost would reduce the model’s practicality as a retrieval system. Therefore, we limited our evaluation to 64 frames and selected 16-frame sampling as the optimal balance between performance and speed.

\subsubsection{Effect of Frame Selection Methods} \label{sec:app_frame_selection_method}

\begin{table*}[!ht]
\resizebox{\textwidth}{!}{
    \begin{tabular}{@{}lccccccccccc@{}}
    \toprule
    Frame Sampling & R@1$\uparrow$ & R@5$\uparrow$ & R@10$\uparrow$ & P@1$\uparrow$ & P@5$\uparrow$ & P@10$\uparrow$ & MRR$\uparrow$ & NDCG$\uparrow$ & MAP$\uparrow$ & MnR$\downarrow$ & MdR$\downarrow$ \\
    \midrule
    MultiCLIP & 9.83 & 44.32 & 70.82 & 88.80 & 80.77 & 65.25 & 0.92 & 75.34 & 86.33 & 22.12 & \textbf{6} \\
    \addlinespace[1mm]
    
    & \multicolumn{11}{c}{\textbf{Without Audio}} \\
    \cmidrule(lr){2-12}
    scene\_detect\footnote{\url{https://www.scenedetect.com/}} & 10.06 & 46.02 & 74.68 & 91.12 & 83.78 & 68.73 & 0.94 & 78.90 & 88.59 & 21.84 & \textbf{6} \\
    uniform (\ours) & \textbf{10.24} & \textbf{46.71} & \textbf{75.76} & \textbf{92.28} & \textbf{85.25} & \textbf{69.73} & \textbf{0.95} & \textbf{80.04} & \textbf{89.42} & \textbf{21.13} & \textbf{6} \\
    
    \addlinespace[1mm]
    
    & \multicolumn{11}{c}{\textbf{With Audio}} \\
    \cmidrule(lr){2-12}

    scene\_detect & \textbf{10.36} & 48.55 & 79.34 & \textbf{93.44} & 87.95 & 72.78 & \textbf{0.95} & 82.99 & 90.98 & 17.58 & \textbf{6} \\
    uniform (\ours) & 10.32 & \textbf{49.00} & \textbf{79.60} & 93.05 & \textbf{88.73} & \textbf{73.09} & \textbf{0.95} & \textbf{83.24} & \textbf{91.20} & \textbf{16.44} & \textbf{6} \\
    \bottomrule
    \end{tabular}
}

\caption{Comprehensive comparison of Different Frame Sampling Methods. In the scene-detect approach, we utilized the \texttt{AdaptiveDetector} with its default settings. For the uniform sampling method, we selected 16 evenly spaced frames, whereas the number of frames in the scene-detect approach varied based on the algorithm's output. Results are reported on the MultiVENT dataset.}
\label{tab:lg_frame_selection_method}
\end{table*}

In \ours, we employed a uniform sampling method for frame selection. However, to compare its effectiveness against a more complex approach, we conducted an ablation study using an adaptive detector from \texttt{PySceneDetector}\footnote{\url{https://www.scenedetect.com/}}. As shown in \cref{tab:lg_frame_selection_method}, uniform sampling achieves the best performance across both variants, demonstrating its robustness regardless of audio availability. Furthermore, the scene-detection algorithm yields a variable number of frames (ranging from 2 to 100), significantly increasing retrieval time by approximately 2.8 times compared to 16-uniform sampling.

\subsubsection{Effect of LLM Sizes} \label{sec:app_llm_sizes}
\begin{table*}[!ht]
\resizebox{\textwidth}{!}{
    \begin{tabular}{@{}lccccccccccc@{}}
    \toprule
    LLM Size & R@1$\uparrow$ & R@5$\uparrow$ & R@10$\uparrow$ & P@1$\uparrow$ & P@5$\uparrow$ & P@10$\uparrow$ & MRR$\uparrow$ & NDCG$\uparrow$ & MAP$\uparrow$ & MnR$\downarrow$ & MdR$\downarrow$ \\
    \midrule
    MultiCLIP & 9.83 & 44.32 & 70.82 & 88.80 & 80.77 & 65.25 & 0.92 & 75.34 & 86.33 & 22.12 & \textbf{6} \\
    \addlinespace[1mm]
    
    & \multicolumn{11}{c}{\textbf{Without Audio}} \\
    \cmidrule(lr){2-12}
    1B & 10.05 & 46.07 & 75.45 & 90.35 & 83.94 & 69.54 & 0.93 & 79.34 & 88.15 & 22.34 & \textbf{6} \\
    3B & 10.18 & 46.70 & 75.50 & 91.51 & \textbf{85.25} & 69.54 & 0.94 & 79.78 & 89.40 & 21.66 & \textbf{6} \\
    8B & 10.21 & 46.60 & 75.02 & \textbf{92.28} & 84.94 & 69.11 & 0.94 & 79.41 & 89.37 & \textbf{19.74} & \textbf{6} \\
    70B (\ours) & \textbf{10.24} & \textbf{46.71} & \textbf{75.76} & \textbf{92.28} & \textbf{85.25} & \textbf{69.73} & \textbf{0.95} & \textbf{80.04} & \textbf{89.42} & 21.13 & \textbf{6} \\
    \addlinespace[1mm]
    
    & \multicolumn{11}{c}{\textbf{With Audio}} \\
    \cmidrule(lr){2-12}
    1B & 10.26 & 48.31 & 78.71 & 92.66 & 87.49 & 72.28 & \textbf{0.95} & 82.50 & 91.23 & 17.64 & \textbf{6} \\
    3B & 10.32 & 48.88 & 79.17 & 93.05 & 88.57 & 72.74 & \textbf{0.95} & 83.03 & \textbf{91.55} & 16.51 & \textbf{6} \\
    8B & \textbf{10.41} & 48.62 & 79.04 & \textbf{93.82} & 88.03 & 72.59 & \textbf{0.95} & 82.91 & 91.28 & 16.79 & \textbf{6} \\
    70B (\ours) & 10.32 & \textbf{49.00} & \textbf{79.60} & 93.05 & \textbf{88.73} & \textbf{73.09} & \textbf{0.95} & \textbf{83.24} & 91.20 & \textbf{16.44} & \textbf{6} \\
    \bottomrule
    \end{tabular}
}

\caption{Comprehensive comparison of different LLM sizes within the Llama-3 family (ranging from 1B to 70B parameters), specifically used for query decomposition and refining frame captions to video caption. Results are reported on the MultiVENT dataset.}
\label{tab:lg_llm_sizes}
\end{table*}

The detailed results for the experiment described in \cref{sec:llm_sizes} are reported on \cref{tab:lg_llm_sizes}.

\subsubsection{Effect of different Rank Fusion} \label{sec:app_rank_fusion}

\begin{table*}[!ht]
\resizebox{\textwidth}{!}{
    \begin{tabular}{@{}lccccccccccc@{}}
    \toprule
    Method & R@1$\uparrow$ & R@5$\uparrow$ & R@10$\uparrow$ & P@1$\uparrow$ & P@5$\uparrow$ & P@10$\uparrow$ & MRR$\uparrow$ & NDCG$\uparrow$ & MAP$\uparrow$ & MnR$\downarrow$ & MdR$\downarrow$ \\
    \midrule
    MultiCLIP & 9.83 & 44.32 & 70.82 & 88.80 & 80.77 & 65.25 & 0.92 & 75.34 & 86.33 & 22.12 & \textbf{6} \\
    \addlinespace[1mm]
    
    & \multicolumn{11}{c}{\textbf{Without Audio}} \\
    \cmidrule(lr){2-12}

    Neg. Exp. Ent. & 9.66 & 39.97 & 60.05 & 87.26 & 73.36 & 55.60 & 0.90 & 66.61 & 83.60 & 68.44 & 7 \\
    RRF & 9.92 & 42.10 & 62.91 & 89.58 & 77.37 & 58.30 & 0.92 & 69.74 & 86.94 & 33.80 & 7 \\
    Max & 9.98 & 45.15 & 71.98 & 90.35 & 82.32 & 66.41 & 0.93 & 76.60 & 87.77 & 29.31 & \textbf{6} \\
    Mean & 10.19 & 45.99 & 73.54 & \textbf{92.28} & 84.09 & 67.88 & 0.94 & 78.37 & \textbf{89.50} & 26.32 & \textbf{6} \\
    Inv. Ent. (\ours) & \textbf{10.24} & \textbf{46.71} & \textbf{75.76} & \textbf{92.28} & \textbf{85.25} & \textbf{69.73} & \textbf{0.95} & \textbf{80.04} & 89.42 & \textbf{21.13} & \textbf{6} \\
    \addlinespace[1mm]
    
    & \multicolumn{11}{c}{\textbf{With Audio}} \\
    \cmidrule(lr){2-12}
    Neg. Exp. Ent. & 10.11 & 43.98 & 67.23 & 90.73 & 80.00 & 61.97 & 0.93 & 73.20 & 87.91 & 59.09 & 7 \\
    RRF & 10.01 & 45.75 & 70.78 & 90.73 & 83.17 & 65.17 & 0.93 & 76.14 & 88.35 & 26.77 & 6 \\
    Max & 9.94 & 46.58 & 76.20 & 89.96 & 84.79 & 70.12 & 0.93 & 79.86 & 88.82 & 22.89 & 6 \\
    Mean & 10.23 & 48.06 & 78.30 & 92.66 & 87.41 & 72.16 & 0.95 & 82.20 & 91.04 & 20.32 & 6 \\
    Inv. Ent. (\ours) & \textbf{10.37} & \textbf{48.42} & \textbf{79.06} & \textbf{93.44} & \textbf{87.80} & \textbf{72.66} & \textbf{0.95} & \textbf{82.89} & \textbf{91.10} & \textbf{17.14} & \textbf{6} \\
    \bottomrule
    \end{tabular}
}

\caption{Full comparison of different aggregation methods for zero-shot rank fusion: Neg. Exp. Ent. (Negative Exponential Entropy), RRF (Reciprocal Rank Fusion), Inv. Ent. (Inverse Entropy), Mean and Max. Results are reported on the MultiVENT dataset.}
\label{tab:lg_agg_methods}
\end{table*}

\begin{table*}[t]
\centering
\begin{tabular}{@{}lr@{\hspace{1em}}r@{\hspace{1em}}r@{\hspace{1em}}r@{}}
\toprule
Aggregation Method & R@10$\uparrow$ & P@10$\uparrow$ & MRR$\uparrow$ & NDCG$\uparrow$ \\
\midrule
Mean over events \& Max over captions & 77.91 & 71.58 & \textbf{0.95} & 81.97 \\
Max over events \& Mean over captions & 0.46 & 0.39 & 0.00 & 0.44 \\
Max over events \& Mean of Top 3 captions & 78.90 & 72.47 & 0.94 & 82.54 \\
Max over event \& Mean over Top 5 captions & 78.91 & 72.47 & 0.94 & 82.50 \\
Mean over Top 3 events \& Max over captions & 79.11 & 72.70 & \textbf{0.95} & 82.91 \\
Mean over Top 3 events \& Mean over Top 3 captions & 78.96 & 72.51 & 0.94 & 82.62 \\
\addlinespace
\textbf{\ours} & \textbf{79.25} & \textbf{72.82} & \textbf{0.95} & \textbf{82.95} \\
\bottomrule
\end{tabular}
\caption{Performance comparison of different aggregation methods across retrieval metrics. \ours, with global maximum, achieves the best performance across all metrics.}
\label{tab:aggregation_results}
\end{table*}

The detailed results for the experiment described in \cref{sec:rank_fusion} are reported on \cref{tab:lg_agg_methods,tab:aggregation_results}. These experiments consider five different approaches for aggregating the scores and performing zero-shot rank fusion. In all approaches, we compute a distribution $P_i$ across all videos, for each of the $i$ score components, as discussed in \cref{sec:rank_fusion}. The specific rank fusion approaches we consider then vary in how we combine each of the five $P_i$ in order to get a single ranking $\hat{S}$ across all videos. Those approaches are:
\begin{description}
\item[Mean Aggregation] We simply average out $P_i$, computing $\hat{S} = \frac{1}{5} \sum_i P_i$.
\item[Max Aggregation] For each video $v$, we maximize across the $P_i$ score for that video: $\hat{S}_v = \max_i P_{i, v}$.
\item[Reciprocal Rank Fusion (RRF)] For each video $v$, we compute the rank of that video in $P_i$, and then sum across those reciprocal ranks. The final rank of $v$ is then $\sum_i ( 1 / Rank(P_{i,v}))$.
\item[Negative Exponential Entropy] We compute $\hat{S} = \sum_i \exp(-H(P_i)) * P_i$, where $H(P_i)$ is the entropy of $P_i$.
\item[Inverse Entropy] This is the approach described in the main paper, in \cref{sec:rank_fusion}.
\end{description}

\subsubsection{Effect of Different Components} \label{sec:app_diff_components}
\begin{table*}[!ht]
\resizebox{\textwidth}{!}{
    \begin{tabular}{@{}lccccccccccc@{}}
    \toprule
    Components & R@1$\uparrow$ & R@5$\uparrow$ & R@10$\uparrow$ & P@1$\uparrow$ & P@5$\uparrow$ & P@10$\uparrow$ & MRR$\uparrow$ & NDCG$\uparrow$ & MAP$\uparrow$ & MnR$\downarrow$ & MdR$\downarrow$ \\
    \midrule
    MultiCLIP & 9.83 & 44.32 & 70.82 & 88.80 & 80.77 & 65.25 & 0.92 & 75.34 & 86.33 & 22.12 & \textbf{6} \\
    \addlinespace[1mm]
    
    & \multicolumn{11}{c}{\textbf{Without Audio}} \\
    \cmidrule(lr){2-12}
    \ours & \textbf{10.24} & \textbf{46.71} & \textbf{75.76} & \textbf{92.28} & \textbf{85.25} & \textbf{69.73} & \textbf{0.95} & \textbf{80.04} & \textbf{89.42} & 21.13 & \textbf{6} \\
    \quad \ablate{Video} & 9.28 & 39.56 & 58.00 & 84.17 & 72.97 & 53.90 & 0.89 & 64.83 & 82.66 & 65.97 & 8 \\
    \quad \ablate{Query} & 10.08 & 45.97 & 74.74 & 91.12 & 83.78 & 68.80 & 0.93 & 78.78 & 87.92 & 21.00 & \textbf{6} \\
    \quad \ablate{Event} & 9.95 & 46.38 & 74.77 & 89.96 & 84.48 & 68.88 & 0.93 & 79.02 & 88.98 & \textbf{18.35} & \textbf{6} \\

    \qquad \ablate{Prequel} & 10.11 & 46.73 & 75.73 & 91.12 & 85.17 & 69.73 & 0.94 & 79.85 & 89.13 & 20.13 & \textbf{6} \\
    \qquad \ablate{ Current} & 10.20 & 46.79 & 75.85 & 91.89 & 85.33 & 69.81 & 0.94 & 80.02 & 89.29 & 20.29 & \textbf{6} \\
    \qquad \ablate{Sequel} & 10.07 & 46.66 & 75.75 & 91.12 & 85.02 & 69.73 & 0.94 & 79.85 & 89.19 & 20.15 & \textbf{6} \\
    \addlinespace[1mm]
    
    & \multicolumn{11}{c}{\textbf{With Audio}} \\
    \cmidrule(lr){2-12}
    \ours & \textbf{10.32} & \textbf{49.00} & \textbf{79.60} & \textbf{93.05} & \textbf{88.73} & \textbf{73.09} & \textbf{0.95} & \textbf{83.24} & \textbf{91.20} & 16.44 & \textbf{6} \\
    \quad \ablate{Video} & 10.16 & 44.61 & 67.74 & 91.51 & 81.16 & 62.43 & 0.93 & 73.96 & 88.97 & 50.96 & \textbf{6} \\
    \quad \ablate{Query} & 10.08 & 47.39 & 78.12 & 91.12 & 85.95 & 71.74 & 0.93 & 81.54 & 89.18 & \textbf{16.32} & \textbf{6} \\
    \quad \ablate{Event} & 10.22 & 48.09 & 77.77 & 91.89 & 87.18 & 71.47 & 0.94 & 81.75 & 90.38 & 16.64 & \textbf{6} \\
    \qquad \ablate{Prequel} & 10.28 & 48.86 & 79.65 & 92.66 & 88.49 & 73.17 & 0.95 & 83.25 & 91.04 & 15.89 & \textbf{6} \\
    \qquad \ablate{Current} & 10.32 & 48.78 & 79.62 & 93.05 & 88.34 & 73.13 & 0.95 & 83.25 & 91.04 & 15.86 & \textbf{6} \\
    \qquad \ablate{Sequel} & 10.32 & 48.91 & 79.34 & 93.05 & 88.57 & 72.86 & 0.95 & 83.09 & 91.25 & 15.81 & \textbf{6} \\
    
    \bottomrule
    \end{tabular}
}

\caption{Full analysis of different components in our method. The top block shows the performance of our method with audio transcripts, while the bottom block represents the performance without audio transcripts. The first row in each block corresponds to our full method with all components included. In subsequent rows, one main component (video, query, Event) is removed at a time to demonstrate its contribution to the overall method. 
While Event consists of Prequel, Current and Sequel, on the bottom three rows, we have removed each of them one at a time to demonstrate the contribution of each. Results are reported on the MultiVENT dataset.}
\label{tab:lg_diff_components}
\end{table*}

The detailed results for the experiment described in \cref{sec:diff_components} are reported on \cref{tab:lg_diff_components}. The fine-grained results for each component of events show a slight reduction in NDCG score (0.1-0.3) if we exclude one of the prequel, sequel, or current generations, if we exclude all, we see a much significant reduction (1-2 points), suggesting strong complementary aspects and a strong aggregate effect.

\subsection{Dataset Statistics}
The detailed dataset statistics are presented on \cref{tab:dataset_specs}.
\begin{table}[t]
\centering
\resizebox{\columnwidth}{!}{
    \begin{tabular}{@{}l|cc|cc@{}}
    \toprule
    Dataset & \# of query & Avg. words & \# of video & Avg. length (sec.)\\
    \midrule
    MSR-VTT-1kA & 995 & 9 & 1000 & 15  \\
    MSVD & 22285 & 8 & 670 & 10 \\
    MultiVENT & 259 & 27 & 2394 & 83 \\
    \bottomrule
    \end{tabular}
}

\caption{Overview of the datasets used in this study. The table summarizes key statistics of the MSR-VTT-1kA \citep{xu2016msr}, MSVD \citep{chen2011building} and MultiVENT \citep{sanders2023multivent} datasets.}
\label{tab:dataset_specs}
\end{table}

\subsection{Implementation Details}
\label{app:implementation}
We used Llama-3\footnote{\href{https://hf.co/meta-llama/Llama-3.3-70B-Instruct}{\texttt{meta-llama/Llama-3.3-70B-Instruct}}} as the LLM, InternVL-2.5\footnote{\href{https://hf.co/OpenGVLab/InternVL2_5-38B}{\texttt{OpenGVLab/InternVL2\_5-38B}}} as the VLM, and Whisper-large-v3 \footnote{\href{https://hf.co/openai/whisper-large-v3}{\texttt{openai/whisper-large-v3}}} as the multilingual ASR model for query decomposition, frame captioning and audio transcription. For embedding computations, we employed ColBERT\footnote{\href{https://hf.co/hltcoe/plaidx-large-zho-tdist-mt5xxl-engeng}{\texttt{hltcoe/plaidx-large}}} for text-to-text similarity and MultiCLIP\footnote{\href{https://hf.co/laion/CLIP-ViT-H-14-frozen-xlm-roberta-large-laion5B-s13B-b90k}{\texttt{laion/Clip-VIT-H14}}} for text-to-video similarity score calculation. Query decomposition and video captioning were performed with a temperature of 0.8 and top-p sampling of 0.95. For frame extraction, we applied uniform sampling with 16 frames. Our evaluation framework is based on torchmetrics\footnote{\url{https://lightning.ai/docs/torchmetrics}}. The data extraction prompts were executed using two A100 GPUs, while the text-to-video retrieval was conducted on an RTX 6000 GPU.

\subsection{Prompts} \label{sec:app_prompts}
In the following sections, we provide all the prompts used for the event decomposition (\cref{app:decompose_query,app:spatial_temporal_event,app:refine_query}), video decomposition (\cref{app:frame_captioner,sec:video_captioner}) and ASR module (\cref{app:refine_asr}).
\subsubsection{Decomposing Query to Prequel, Current, and Sequel Events} \label{app:decompose_query}

The following prompt was utilized to extract prequel, current, and sequel events from the query, with minimal modifications to maintain generalizability across all cases.

\prompt[
{\color{WildStrawberry}Prequel} / {\color{Cerulean} Current} / {\color{Green} Sequel}
]{
You are given a video search query. Assume the video search query is an ongoing event. You have to {\color{WildStrawberry}find out what previous events could lead to this given event} / {\color{Cerulean}decompose the ongoing event in multiple simple events} / {\color{Green} find out what events can be the outcome of this ongoing event}.The extracted events should be concrete enough to be visualized in a video.
\vspace{2mm}

Here are some guidelines: 

- Extract events that could be visualized in a video.

- Before extracting an event, consider whether the event text could serve as a reasonable caption for a frame of a video associated with the query.

- Give specific events related to the query, use your previous WORLD KNOWLEDGE to extract events that are likely to be seen in a video. For example, if the query is "2019 Christmas Eve Protests in Hong Kong", you might return "Protesters marching in the streets", but not "Protestors marching in snow" because it is not likely to have snow in Hong Kong, though its pretty common in other places. 

- First, explain your reasoning in 1-2 sentences about which events it makes sense to extract. Then, provide a numbered list with the events (anywhere from one to five event; use your best judgement to determine how many to extract)

- Structure your response into two sections, ""EXPLANATION:"" and then ""EVENTS:"". The events sections should contain nothing but the events. 
\vspace{2mm}

Here is the query I'd like you to extract events from: \promptvar{query}
}

\subsubsection{Extracting Spatial, Temporal, and Event Data} \label{app:spatial_temporal_event}
The following three prompts were utilized to extract the primary event, location, and time of the events from the query, respectively. The only input for these prompts is the original query. In cases where the query does not mention a specific time or location, the LLM was instructed to output ``NOT AVAILABLE.''

\prompt[Extract Primary Event]{
You are given a video search query. Your task is to extract the primary event(s) being referenced or discussed in the query.  
\vspace{2mm}

Here are some guidelines:

- Identify the main event(s) that the query is about, whether explicitly stated or implied

- If multiple events are mentioned, list all of them

- If no clear event can be identified from the query, output "NOT AVAILABLE" in the EVENTS section

- First, explain your reasoning in 1-2 sentences about how you identified the event(s) from the query

- Structure your response into two sections, "EXPLANATION:" and then "EVENTS:"

- The events section should contain either the identified event(s) in a numbered list, or "NOT AVAILABLE" if no event can be determined

- Don't include temporal or spatial information in this event, just focus on the primary event(s) being discussed or referenced
\vspace{2mm}

Here is the query I'd like you to extract events from: \promptvar{query}
}

\prompt[Extract Spatial Information]{
You are given a video search query. Assume the video search query is an ongoing event. Your task is to extract any locations mentioned in the query.
\vspace{2mm}

Here are some guidelines:

- Extract any location information mentioned in the query, including specific cities, towns, countries, regions, areas, neighborhoods, or other geographical references

- If no location information is found in the query, output "NOT AVAILABLE" in the LOCATION INFORMATION section

- First, explain your reasoning in 1-2 sentences about what location information appears in the query (or lack thereof) and why you're extracting it

- Structure your response into two sections, "EXPLANATION:" and then "LOCATION INFORMATION:"

- The location information section should contain either the extracted location information in one line, or "NOT AVAILABLE" if no location information is provided.

- Output in the following format: CITY, COUNTRY, REGION, AREA, NEIGHBORHOOD, etc. (e.g. New York, USA, Europe, Downtown). If one of the fields is not available, you can leave it out.
\vspace{2mm}

Here is the query I'd like you to extract location information from: \promptvar{query}
}

\prompt[Extract Temporal Informaton]{
You are given a video search query. Assume the video search query is an ongoing event. Your task is to extract any dates, times, or years mentioned in the query.
\vspace{2mm}

Here are some guidelines:

- Extract any temporal information mentioned in the query, including specific dates, times of day, years, seasons, or other time-related references

- If no temporal information is found in the query, output "NOT AVAILABLE" in the TEMPORAL INFORMATION section

- First, explain your reasoning in 1-2 sentences about what temporal information appears in the query (or lack thereof) and why you're extracting it

- Structure your response into two sections, "EXPLANATION:" and then "TEMPORAL INFORMATION:"

- The temporal information section should contain either the extracted date/time information in one line, or "NOT AVAILABLE" if no temporal information is provided.

- Output in the following format: DAY MONTH YEAR, TIME, SEASON, etc. (e.g. 1 January 2022, 3:00 PM, Summer). If one of the fields is not available, you can leave it out.
\vspace{2mm}

Here is the query I'd like you to extract temporal information from: \promptvar{query}
}

\subsubsection{Refining Query} \label{app:refine_query}

After extracting information from the previous prompts, the following prompt was employed to refine the prequel, current, and sequel events by incorporating the extracted event, temporal, and spatial details. The goal was to construct natural, coherent, human-like search queries while preserving all relevant information.

\prompt[
Refine Query
]{
You will receive a query along with optional time, location, and main event information. Your task is to integrate this information naturally into a refined, human-like search query.
\vspace{2mm}

Here are some guidelines:

- Take the original query and incorporate any provided time, location, and event details seamlessly

- If any of these elements (time, location, event) are not provided, work with just the available information

- The refined query should sound natural, as if someone is searching with that query in Google, YouTube, etc.

- The base query information must always be preserved in the refined version

- First, explain your reasoning in 1-2 sentences about which events it makes sense to extract. Then, provide the refine query in one line

- Structure your response into two sections, ""EXPLANATION:"" and then ""REFINED QUERY:"". The Refined Query section should contain nothing but the refined query in one line.
\vspace{2mm}

Base Query: \promptvar{prequel/Current/sequel}

Event: \promptvar{event}

Place: \promptvar{place}

Time: \promptvar{time}
}

\subsubsection{Frame Captioner} \label{app:frame_captioner}
The following prompt was used to generate captions for each video frame. Unlike other prompts, a VLM was utilized to handle the multimodal inputs effectively. As illustrated in \cref{fig:main_with_asr}, we implemented two variants: (1) ASR and (2) Non-ASR. The {\color{YellowOrange} Orange} texts were excluded in the prompt for the Non-ASR variant.

\prompt[
Frame Caption
]{
You are provided with the description of the previous frames, the current frame (image) of a single video{\color{YellowOrange}, and the original transcript of the audio (ASR) from the entire video}. Your task is to describe the main event or activity depicted in the current frame.
\vspace{2mm}

\# Instructions:

{\color{YellowOrange}- Use the original ASR to enhance your understanding of the video context and inform your description of the current frame.}

- Focus on identifying and detailing the key actions, participants, and context of the event depicted in the current frame.

- Maintain continuity with the previous frame’s description, connecting the current frame to the overall flow of the video.

- Ignore unrelated visual elements like colors or shapes unless they are critical to understanding the event.

- Use any visible text in the image to enhance your interpretation of the frame.
\vspace{2mm}

Only output the current frame description, integrating insights from {\color{YellowOrange}the original ASR and} previous frame description. Don't output any additional commentary or content.
\vspace{2mm}

{\color{YellowOrange}\# Original ASR Transcript (Entire Video):

\{\{ASR\}\}
}
\vspace{2mm}

\# Previous Frame Description:

\promptvar{prev\_caption}
\vspace{2mm}

\# Frame:

\promptvar{frame}
\vspace{2mm}

\# Current Frame Description:

}

\subsubsection{Full Video Captioner} \label{sec:video_captioner}
The following prompt generated a summarized caption for the entire video. Given the high computational requirements and the limited capability of VLMs in handling videos, we utilized an LLM to summarize all the captions extracted in the previous step. The full refined audio transcript was also included in the ASR variant to provide additional context. As before, the {\color{YellowOrange} Orange} text elements were excluded from the prompt in the Non-ASR variant.

\prompt[
Video Caption using Frame Captions
]{
You will receive individual descriptions of consecutive frames from a single video{\color{YellowOrange}, along with the original transcript of the audio (ASR) from the entire video}. Your task is to provide a detailed description of the whole video, summarizing the key actions, participants, and context of the event depicted in the video.
\vspace{2mm}

\# Instructions:

{\color{YellowOrange}- Use the original ASR to enhance your understanding of the video context and inform your description of the whole video.}

- Focus only on relevant aspects, such as what is happening, who is involved, and the overall context or setting.

- Ignore unrelated visual details like colors, shapes, or other non-essential information.

- Your summary should concisely and accurately capture the essence of the event, integrating insights from the frame descriptions{\color{YellowOrange}, the Original ASR and the ASR summary}.
\vspace{2mm}

Only output the video summary, without any additional commentary or content.
\vspace{2mm}

\# Frame Descriptions:

\#\# Frame {\color{cyan}[i]} Description

\promptvar{frame\_i\_description}
\vspace{2mm}

{\color{YellowOrange}\# Original ASR Transcript (Entire Video):

\{\{ASR\}\}
}
\vspace{2mm}

\# Summary:

}

\subsubsection{Refining Audio Transcripts} \label{app:refine_asr}
As illustrated in \cref{fig:main_with_asr} (left orange box), we utilized an LLM to refine the translations generated by the multilingual-ASR and a translation-specific model. The following prompt was employed to refine the audio transcript using three inputs: (1) the original language transcript from ASR, (2) the translated transcript from the multilingual-ASR, and (3) the translated transcript from the translator.

\prompt[
Refine Audio Transcript
]{
You are an expert in language translation and text refinement. Your task is to produce a highly accurate and natural-sounding translation by analyzing the original transcript and two provided translations.
\vspace{2mm}

\# Instructions:

- Combine the strengths of both translations and use the original transcript as a reference to improve accuracy and contextual meaning.

- Prioritize clarity, fluency, and maintaining the original intent and tone of the text.

- Where translations differ significantly, cross-check with the original transcript for fidelity.

- Ensure the final output is free from any awkward phrasing or mistranslation.

- If the transcripts can't be refined with reasonable accuracy, just output "Not Available".
	
\# Transcripts:

\#\# Original Transcript:

\promptvar{original\_language\_asr}
\vspace{2mm}

\#\# Translation 1:

\promptvar{whisper\_translated\_asr}
\vspace{2mm}

\#\# Translation 2:

\promptvar{nllb\_translated\_asr}
\vspace{2mm}

Output the refined translation only, without additional notes or explanations.
\vspace{2mm}

\# Refined Translation:

}

\subsection{Use of AI Assistance}
The authors have used GitHub copilot \footnote{\url{https://github.com/features/copilot}} during development and ChatGPT\footnote{\url{https://chatgpt.com/}} for proofreading and polishing the final writing. Content provided to those tools were original to the authors.

\end{document}